\documentclass{article}

 \usepackage[dblblindworkshop, final]{neurips_2025}
\workshoptitle{Structured Probabilistic Inference \& Generative
Modeling workshop of NeurIPS 2025}

\usepackage[utf8]{inputenc} 
\usepackage[T1]{fontenc}    
\usepackage{hyperref}       
\usepackage{url}            
\usepackage{booktabs}       
\usepackage{amsfonts}       
\usepackage{nicefrac}       
\usepackage{microtype}      
\usepackage{xcolor}         

\usepackage{xspace}
\usepackage{extra_packages}
\title{Enhancing Diffusion Model Guidance through Calibration and  Regularization} 
\newtheorem{assumption}{Assumption}

\author{%
  Seyed Alireza Javid \\
  UC San Diego\\
  \texttt{sajavid@ucsd.edu}
  \And
  Amirhossein Bagheri \\ 
  Politecnico di Milano\\
  \texttt{amirhossein.bagheri@mail.polimi.it}
  \And
  Nuria Gonz\'alez-Prelcic \\
  UC San Diego\\
  \texttt{ngprelcic@ucsd.edu}
}
\begin{document}

\maketitle

\begin{abstract}
Classifier-guided diffusion models have emerged as a powerful approach for conditional image generation, but they suffer from overconfident predictions during early denoising steps, causing the guidance gradient to vanish. This paper presents two complementary contributions to enhance classifier-guided diffusion models. First, we introduce a differentiable Smooth Expected Calibration Error (Smooth ECE) loss that improves classifier calibration with minimal fine-tuning, achieving approximately a 3\% improvement in Fréchet Inception Distance (FID) scores. Second, we propose enhanced sampling guidance methods that operate on off-the-shelf classifiers without requiring retraining. Our approach includes: (1) tilted sampling that leverages batch-level information to control outlier influence, (2) adaptive entropy-regularized sampling to maintain diversity, and (3) a novel divergence-regularized sampling method that adds a class-aware, mode-covering correction that strengthens movement toward the target class while maintaining exploration. Theoretical analysis reveals that our methods effectively combine enhanced
target direction guidance with controlled diversity exploration, mitigating gradient vanishing. Experimental results on ImageNet demonstrate that our best divergence-guided sampling achieves an FID of 2.13 while maintaining competitive precision and recall metrics. Our methods provide a practical solution for improving conditional generation quality without the computational overhead of classifier and diffusion model retraining. \textit{Code is available at:} \url{https://github.com/ajavid34/guided-info-diffusion}.
\end{abstract}

\section{Introduction}
Denoising diffusion probabilistic models (DDPMs) have achieved state-of-the-art results in unconditional image generation, producing high-quality and diverse images by progressively reversing a noising process \cite{ho2020denoising, nichol2021improved, song2020denoising}. These models are built on a solid probabilistic foundation, which allows for stable training and scalability across a wide range of datasets. A key strength of DDPMs is their adaptability: by incorporating external information like class labels or text embeddings, they can be extended to conditional image generation.

Conditional image generation enables the creation of images that adhere to specific constraints, such as class labels or textual descriptions \cite{ramesh2021zero, rombach2022high, liu2024sora}. DDPMs have proven to be highly effective in this area, particularly for generating class-conditioned images of exceptional quality \cite{ho2020denoising, dhariwal2021diffusion, nichol2021improved}. The theoretical underpinnings of diffusion models, similar to score-based generative models, allow for this conditional extension through Bayes' theorem without the need for retraining the entire model \cite{song2023consistency, song2019generative, song2020score}. One of the most effective techniques for conditional generation involves using an independent, noise-aware classifier to guide the reverse diffusion process \cite{dhariwal2021diffusion, sohl2015deep, ma2023elucidating}. This classifier directs the generation process toward the desired output by estimating the gradient of the log-probability of the target label with respect to the noisy input, all without requiring retraining of the generative model itself.

Despite these advances, a critical limitation of classifier-guided diffusion models is that they suffer from overconfident predictions during early denoising steps, causing the guidance gradient to vanish \cite{ma2023elucidating, zheng2022entropy}. Our main contributions to this paper are as follows:
\begin{itemize}
\item {
We propose a calibration-aware classifier finetuning loss based on the Smooth Expected Calibration Error (Smooth ECE) \cite{naeini2015obtaining}. This loss improves calibration and achieves better downstream performance while introducing only minimal fine-tuning overhead.}
\item {
We introduce an entropy-regularized guidance method that prevents premature overconfidence, promoting diversity during sampling while preserving class relevance.}

\item {
We propose a divergence regularized guidance method, which enhances mode coverage and preserves fidelity to the target class distribution.}

\item {
We design a tilted-loss-based batch-aware sampling strategy, which leverages information across generated samples to balance quality and diversity without additional complexity.}


\end{itemize}
We conduct experimental analysis on ImageNet $128\times128$, demonstrating that our proposed methods outperform existing classifier-guided diffusion approaches in terms of Fréchet inception distance (FID) \cite{heusel2017gans}, precision, and recall.



\section{Classifier design}

As introduced by \cite{ma2023elucidating}, the ECE (details in Appendix \ref{ECE-Smooth})  has a direct connection to the FID. Motivated by this connection, we define a differentiable Huber-like calibration loss termed \textbf{Smooth ECE}, which operates over confidence bins and applies a smooth absolute error at the individual sample level. The loss is formulated as
\begin{equation}
    \mathcal{L}_{\mathrm{ECE}} 
    = \frac{1}{n} \sum_{b=1}^{B} \sum_{i: \hat{p}^{(i)} \in \mathcal{B}_b} 
    \sqrt{ \left( \hat{p}^{(i)} - a^{(i)} \right)^2 + \beta }.
\end{equation}
where $\hat{p}^{(i)} = \max_{y} p_\phi(y \mid x^{(i)})$ is the predicted confidence for sample $i$, and $a^{(i)} = \mathbb{I}[\hat{y}^{(i)} = y^{(i)}]$ is the correctness of the prediction, where $\hat{y}^{(i)} = \arg\max_c p_\phi(c \mid x^{(i)})$. The term $\beta > 0$ is a smoothing constant that ensures differentiability. The interval $\mathcal{B}_b = \left( \frac{b-1}{B}, \frac{b}{B} \right]$ defines the $b$-th confidence bin, and each sample contributes only to the bin corresponding to its confidence level. The loss is computed per sample and aggregated over all bins to form the final scalar value. The relationship between smooth ECE and Huber loss is explained in Appendix \ref{ECE-Smooth}.

\begin{remark}
    Our method differs from Meta-Calibration \cite{bohdal2023metacalibration} by directly introducing smoothness into the ECE loss via a Huber-like function, avoiding the need for soft binning or differentiable ranking. Unlike their meta-learning framework, we apply our smooth ECE loss directly during training without outer-loop optimization. This results in a simpler, more efficient, and plug-and-play approach.
\end{remark}



\section{Sampling guidance}
We proposed a calibration-aware classifier design that leverages the ECE. However, fine-tuning classifiers can be costly and time-consuming. In what follows, we focus on off-the-shelf classifier-guidance methods that operate directly on existing checkpoints and require no additional fine-tuning. Since off-the-shelf classifiers are typically not robust to Gaussian noise and are not time-dependent \cite{ma2023elucidating}, we select $\widehat{x}_0(t)=\left(\widehat{x}_t-\sqrt{1-\bar{\alpha}_t} \epsilon_\theta\left(\widehat{x}_t, t\right)\right) / \sqrt{\bar{\alpha}_t}$ as the classifier input.

\subsection{Regularized sampling} \label{rsample}
The sampling guidance from the classifier during the denoising steps of a diffusion model is helpful for increasing both the diversity and quality of the generated image \cite{dhariwal2021diffusion}. However, a critical flaw emerges in this process. Even when an image is only partially generated and still lacks fine-grained features, a standard classifier can often classify it with excessively high confidence \cite{zheng2022entropy}. This causes the classifier's predicted distribution for the noisy image to prematurely converge to a one-hot distribution. As a result, the conditional gradient guidance vanishes early in the process, and the conditional generation degrades into a less effective unconditional one for the remaining steps.

\begin{figure}[!h]
    \centering
    \includegraphics[width=.66\linewidth]{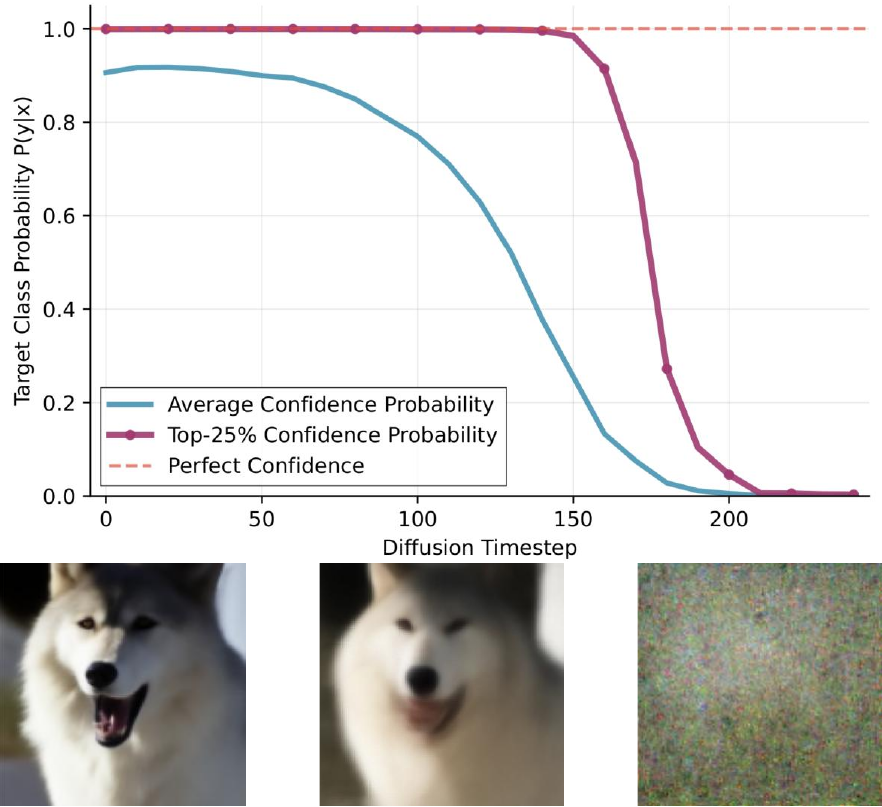}
    \caption{
    The visualization of the denoising sampling process. Overconfidence in classifier-guided diffusion sampling without regularization is mostly effective in the last steps where the probability is close to one.}
    \label{fig:over}
\end{figure}

To address this limitation, \cite{zheng2022entropy} proposed entropy-constrained training and scalar weight based entropy on the classifier guidance. While they show some improvement compared to the baseline, their scheme requires costly training of a classifier from scratch and cannot be generalized to off-the-shelf classifiers.
As Figure~\ref{fig:over} shows, the top 25\% most confident samples maintain near-perfect confidence throughout most of the sampling process, demonstrating the overconfidence problem that leads to less effective guidance. While the average confidence increases more gradually, we can still see the overconfidence problem for most of the denoising process. This motivates our analysis of two general methods for more robust classifier guidance.
\subsubsection{Entropy-regularized sampling}
\cite{ma2023elucidating} demonstrated that the classifier guidance can be interpreted as the gradient of the joint $E_{\tau_1}(x, y)$ and marginal energy $E_{\tau_2}(x)$ difference as follows:
\begin{equation}
    \begin{aligned}
    \log p_{\tau_1, \tau_2}(y \mid x)&=  \log \frac{\exp \left(\tau_1 f_y(x)\right)}{\sum_{i=1}^N \exp \left(\tau_2 f_i(x)\right)}=\tau_1 f_y(x)-\log \sum_{i=1}^N \exp \left(\tau_2 f_i(x)\right)\\
    :&=-E_{\tau_1}(x, y)+E_{\tau_2}(x),
        \end{aligned}
\end{equation}
{\fontsize{8.5}{10}\selectfont
\begin{equation}\label{energy-eq1}
    \begin{aligned}
    \nabla_x \log p_{\tau_1, \tau_2}(y \mid x)&=\nabla_x\left(\tau_1 f_y(x)-\log \left(\sum_{i=1}^N \exp \left(\tau_2 f_i(x)\right)\right)\right):=-\nabla_x E_{\tau_1}(x, y)+\nabla_x E_{\tau_2}(x), 
            \end{aligned}
\end{equation}
}
where $\tau_1$ and $\tau_2$ represent joint and marginal temperatures, respectively. Based on the discussion in Section~\ref{rsample}, we rewrite the regularized guidance as
\begin{equation}
    \mathcal{S}_{\text{entropy}}(x, y) := \log p_{\tau_1,\tau_2}(y|x) + \lambda_t H(p(\cdot|x)),
\end{equation}
where $H(p(\cdot|x)) = -\sum_{i=1}^N p(i|x) \log p(i|x)$ is the entropy of the distribution, $p(i|x) = \frac{\exp( f_i(x))}{\sum_{j=1}^N \exp( f_j(x))}$, and $\lambda (t)\geq 0$ is the adaptive entropy regularization weight. 
\begin{proposition}
\label{pr1}
The gradient of the entropy-regularized score can be written as
{\fontsize{8.5}{10}\selectfont
\begin{align}
\nabla_x \mathcal{S}_{\text{entropy}}(x, y) = &\tau_1 \nabla_x f_y(x) - \tau_2 \sum_{i=1}^N p_{\tau_2}(i|x) \nabla_x f_i(x)\\
&- \lambda_t \sum_{i=1}^N p(i|x)[\log p(i|x) + 1] \left[\nabla_x f_i(x) - \sum_{j=1}^N p(j|x) \nabla_x f_j(x)\right].
\end{align}
}

\end{proposition}

We observe that the entropy gradient promotes diversity by driving the distribution away from deterministic solutions, which balances exploration in high-confidence regions. Specifically, the entropy term $\nabla_xH(p(\cdot|x))$ contributes a weighted sum of gradients from all classes, where each class's contribution is modulated by $ p(i|x) \left[\log p(i|x) + 1\right]$. This encourages the model to maintain uncertainty with a regularization during sampling, effectively balancing exploitation of the target class with exploration of the probability landscape.

However, a fundamental limitation of this approach is that the entropy regularization is class-agnostic. The sum over the gradient of all classes in the regularization term which is independent of the target class, might overshoot the gradient. This indiscriminate diversity enhancement can lead to suboptimal gradient updates, where the model is encouraged to explore directions that increase entropy but potentially move away from the target class manifold. Consequently, while the method successfully prevents overconfidence, it may compromise the fidelity and class-consistency of the generated images, particularly in later denoising steps where precise class alignment becomes crucial. 

\subsubsection{Divergence-regularized sampling}
Based on the mentioned limitation of the entropy approach, we introduce the $f$-divergence regularized guidance as 
\begin{equation}
    \mathcal{S}_{\mathcal{D}}(x, y) := \log p_{\tau_1,\tau_2}(y|x) - \alpha D_{f}(q_y(\cdot) \| p(\cdot|x)),
\label{eq:f_divergence_general}
\end{equation}
where $D_f(q \| p) = \sum_i q(i) f\left(\frac{p(i)}{q(i)}\right)$ is an $f$-divergence with generator function $f: \mathbb{R}_+ \to \mathbb{R}$, $q_y(\cdot)$ is a target distribution (discussed further in Appendix~\ref{addexdiv}), and $\alpha > 0$ is the divergence weight. We use $q_y(i) = (1-\epsilon)\frac{1}{N} + \epsilon\mathbb{I}_{i=y}$ with $\epsilon=0.1$, where the uniform component $(1-\epsilon)/N > 0$ ensures $q_y(i) > 0$ for all $i$. This smoothing is essential: if $q_y$ were one-hot, the divergence $D_f(q_y\|p)$ would be infinite unless $p$ is also one-hot (which never occurs during sampling). The uniform component prevents divergence blow-up while $\epsilon$ provides target class emphasis. Different choices of $f$ lead to different divergences with distinct theoretical properties and practical implications for guidance.
\begin{assumption}
\label{assump:divergence}
The generator function $f: \mathbb{R}_+ \to \mathbb{R}$ satisfies:
\begin{enumerate}
    \item $f$ is differentiable on $(0,\infty)$,
    \item $f(1) = 0$,
    \item The target distribution satisfies $q_y(i) > 0$ for all $i$ to ensure finite divergence.
\end{enumerate}
\end{assumption}

\begin{proposition}
\label{prop:general_f_div}
The gradient of the $f$-divergence regularized score is
\begin{align}
\nabla_x \mathcal{S}_{\mathcal{D}}(x,y)
&= \tau_1 \nabla_x f_y(x)
 - \tau_2 \!\sum_{i=1}^N p_{\tau_2}(i|x)\,\nabla_x f_i(x) \nonumber\\
&\quad - \alpha \!\sum_{i=1}^N 
   w_f(q_y(i),p(i|x))\,
   g_i(x),
\label{eq:general_f_gradient}
\end{align}
where $p(i|x) = \frac{\exp(f_i(x))}{\sum_j \exp(f_j(x))}$,
$g_i(x)= \nabla_x f_i(x) - \sum_{j=1}^N p(j|x)\,\nabla_x f_j(x)$, $p_{\tau_2}(i|x) = \frac{\exp(\tau_2f_i(x))}{\sum_j \exp(\tau_2f_j(x))}$  and 
$w_f(q,p)=p\,f'\!\left(\frac{p}{q}\right)$.
\end{proposition}

We now examine three specific $f$-divergences, each offering distinct theoretical properties and practical trade-offs for diffusion guidance.
\subsubsection{Reverse KL Divergence: Mode-Covering Guidance}
\label{sec:rkl}

For reverse Kullback-Leibler divergence with $f(t) = -\log(t)$, we have $f'(t) = -1/t$, which gives $w_f(q, p) = -q$. This yields a particularly interpretable gradient form.

\begin{corollary}[Reverse KL gradient]
\label{cor:rkl_gradient}
The gradient of the reverse KL divergence regularized score is:{\fontsize{8.5}{10}\selectfont
\begin{equation}
\begin{aligned}
\nabla_x \mathcal{S}_{\text{RKL}}(x, y) =& \tau_1 \nabla_x f_y(x) - \tau_2 \sum_{i=1}^N p_{\tau_2}(i|x) \nabla_x f_i(x) \\&+ \alpha \sum_{i=1}^N q_y(i) \left[\nabla_x f_i(x) - \sum_{j=1}^N p(j|x) \nabla_x f_j(x)\right].
\end{aligned}
\label{eq:rkl_gradient}
\end{equation}}
\end{corollary}

Reverse KL divergence exhibits two fundamental properties that are particularly advantageous for our guidance objective \citep{polyanskiy2025information, bishop2006pattern}. First, its \emph{mode-covering} behavior ensures that the model distribution $p(\cdot|x)$ maintains non-zero probability mass wherever the target distribution $q_y(\cdot)$ has support, effectively preventing the exclusion of any viable modes during sampling. Second, its \emph{zero-avoiding} characteristic imposes a severe penalty when $p(\cdot|x) \to 0$ while $q_y(\cdot) \geq 0$, which mathematically manifests as the divergence approaching infinity. This asymmetric penalty structure is crucial for diversity preservation. In the context of diffusion guidance, these properties ensure that the generated samples maintain coverage over all classes in the target distribution.

To gain deeper insight into the guidance mechanism, we decompose the reverse KL gradient into interpretable components.

\begin{lemma}[Reverse KL decomposition]
\label{lem:rkl_decomposition}
The negative of reverse KL gradient can be decomposed as:{\fontsize{8.5}{10}\selectfont
\begin{equation}
-\nabla_x D_{\text{KL}}(q_y \| p(\cdot|x)) =  \underbrace{\sum_{i=1}^N q_y(i) \nabla_x f_i(x)}_{\text{Target direction}} - \underbrace{\sum_{j=1}^N p(j|x) \nabla_x f_j(x)}_{\text{Current direction}}.
\label{eq:rkl_decomposition}
\end{equation}}
\end{lemma}

This decomposition reveals that the reverse KL regularization pulls the sample toward the target class distribution $q_y$ while simultaneously pushing away from the current prediction $p(\cdot|x)$. This dual mechanism prevents premature convergence to overconfident predictions.

To understand the guidance direction more concretely, we conduct a closed-form analysis in mixed-Gaussian scenarios, which provides insight into how the method operates on structured data manifolds.

\begin{proposition}[Gaussian mixture analysis]
\label{prop:gaussian_rkl}
Let $X \sim \mathcal{P}$ be a random variable defined on $\mathbb{R}^d$, with density function
$f(x) = \sum_{k=1}^K b_k f_k(x)$, where $f_k(x)$ is a normal density with mean $\mu_k$ and covariance matrix
$\Sigma_k$, and $b_k > 0$ with $\sum_{k=1}^K b_k = 1$.
Define the posterior as $w_k(x) := p(k|x)=\frac{b_k f_k(x)}{\sum_{j=1}^K b_j f_j(x)}$. Then:
\begin{equation}
\nabla_x \mathcal{S}_{\text{RKL}}(x, y) = \sum_{k=1}^K f_k(x)\Sigma_k^{-1}(\mu_k - x)\, \Gamma_k(x, y, \alpha, \epsilon),
\label{eq:gaussian_rkl}
\end{equation}
where the weight function is:
\begin{equation}
\Gamma_k(x, y, \alpha, \epsilon)
= \tau_1 \mathbb{I}_{k=y}
- \tau_2\, p_{\tau_2}(k|x)
+ \alpha\big(q_y(k) - w_k(x)\big),
\label{eq:gamma_weight}
\end{equation}
and 
$
w_k(x)
$ is the posterior distribution. Here $p_{\tau_2}(k|x)$ is the tempered posterior over components, e.g.
$p_{\tau_2}(k|x)\propto \exp(\tau_2\,\ell_k(x))$ with $\ell_k(x)=\log b_k+\log f_k(x)$,
and $\epsilon$ enters only through $q_y$ when you choose a smoothed target.
\end{proposition}

\begin{remark}
\label{rem:identity_covariance}
When all covariances are identity matrices ($\Sigma_k = \sigma^2 I$) and the target distribution follows $q_y(i) = (1-\epsilon)\frac{1}{N} + \epsilon \mathbb{I}_{i=y}$, the guidance is proportional to:
\begin{equation}
\frac{1}{\sigma^2}\left[\underbrace{(\tau_1 + \alpha\epsilon)(\mu_y - x)}_{\text{Enhanced target direction}} + \alpha\frac{1-\epsilon}{K}\underbrace{\sum_{k \neq y}(\mu_k - x)}_{\text{Diversity directions}} - \underbrace{\sum_{k=1}^K (\tau_2 p_{\tau_2}(k|x) + \alpha p(k|x))(\mu_k - x)}_{\text{Current distribution pull}}\right].
\label{eq:identity_rkl}
\end{equation}
The first term pulls toward the target class center with amplified strength, the second term uniformly pulls toward all non-target class centers to maintain mode coverage, and the last term acts as an adaptive correction that prevents overshooting and maintains stability.
\end{remark}

\begin{figure}[!h]
    \centering
    \includegraphics[width=.85\linewidth]{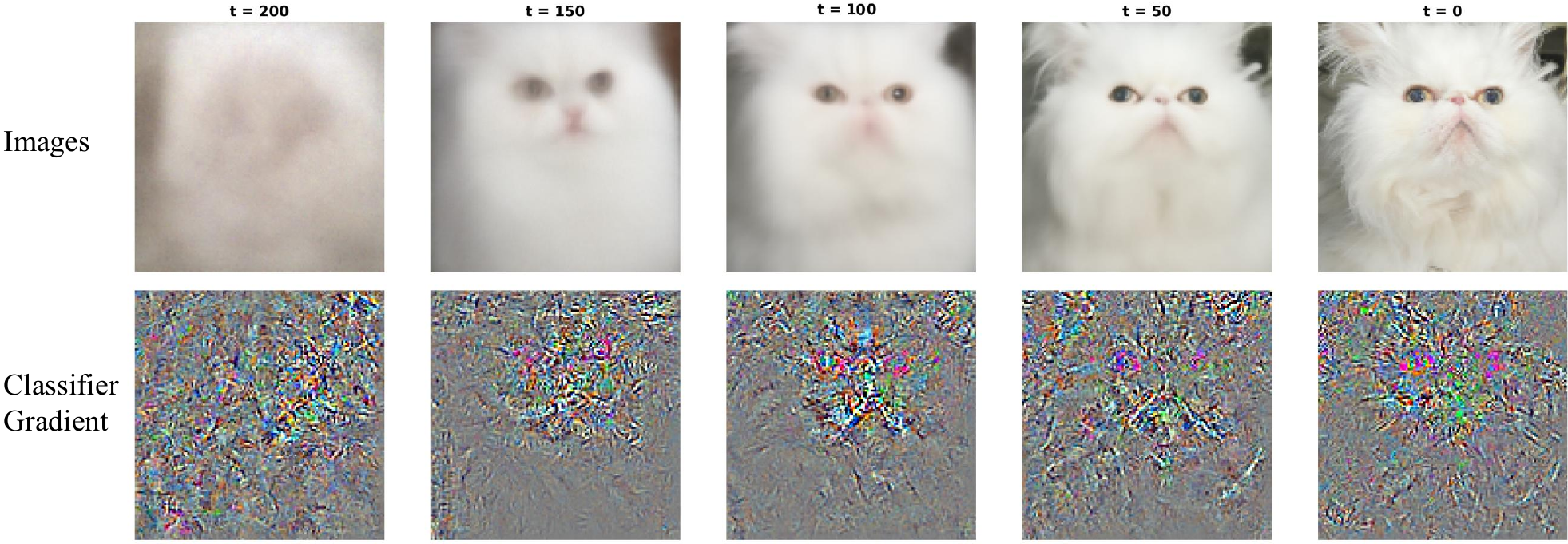}
    \caption{The visualization of intermediate sampling pictures and classifier gradient figures.}
    \label{fig:grad}
\end{figure}

Figure~\ref{fig:grad} demonstrates that throughout the entire sampling process, the gradient maps maintain significant activity across multiple regions, never collapsing to a single point. This confirms the mode-covering property of reverse KL divergence. Additional visualizations are provided in Appendix~\ref{addexp}.
\subsubsection{Forward KL Divergence: Mode-Seeking Guidance}
\label{sec:fkl}

In contrast to reverse KL, forward KL divergence with $f(t) = t \log(t)$ exhibits mode-seeking behavior, penalizing $p(\cdot|x)$ for placing mass where $q_y(\cdot)$ has little support. For this divergence, we have $w_f(q, p) = p[\log(p/q) + 1]$.
\begin{corollary}[Forward KL gradient]
\label{cor:fkl_gradient}
The gradient of the forward KL divergence regularized score is:
\begin{equation}
\begin{aligned}
\nabla_x S_{FKL}(x,y) &= \tau_1\nabla_x f_y(x) - \tau_2 \sum_{i=1}^N p_{\tau_2}(i|x)\nabla_x f_i(x) \nonumber\\
&\quad - \alpha \sum_{i=1}^N p(i|x)\left[\log\frac{p(i|x)}{q_y(i)} + 1\right] \nonumber\\
&\quad\quad \times \left[\nabla_x f_i(x) - \sum_{j=1}^N p(j|x)\nabla_x f_j(x)\right].
\end{aligned}
\end{equation}
\label{eq:fkl_gradient}
\end{corollary}

This formulation prioritizes precision over coverage. When $q_y$ is concentrated on a single mode, forward KL strongly penalizes $p(\cdot|x)$ for exploring other modes, potentially producing sharper but less diverse samples compared to reverse KL. The logarithmic weighting term $\log(q_y(i)/p(i|x))$ becomes increasingly negative for modes where $p(\cdot|x)$ exceeds $q_y$, creating a strong repulsive force away from non-target regions. The numerical results also confirm this in Section~\ref{exp} as this method has the highest precision and lowest recall. 
\subsubsection{Jensen-Shannon Divergence: Balanced Guidance}
\label{sec:js}

Jensen-Shannon divergence offers a symmetric alternative that balances mode-seeking and mode-covering behaviors through an implicit mixture distribution $m = \frac{1}{2}(q_y + p(\cdot|x))$. It can be expressed as:
\begin{equation}
D_{\text{JS}}(q_y \| p(\cdot|x)) = \frac{1}{2}D_{\text{KL}}(q_y \| m) + \frac{1}{2}D_{\text{KL}}(p(\cdot|x) \| m),
\label{eq:js_definition}
\end{equation}
where $m(i) = \frac{1}{2}(q_y(i) + p(i|x))$.
\begin{corollary}[Jensen-Shannon gradient]
\label{cor:js_gradient}
The gradient of the Jensen-Shannon divergence regularized score is:
\begin{equation}
\begin{aligned}
\nabla_x S_{JS}(x,y) &= \tau_1\nabla_x f_y(x) - \tau_2 \sum_{i=1}^N p_{\tau_2}(i|x)\nabla_x f_i(x) \nonumber\\
&\quad - \alpha \sum_{i=1}^N p(i|x)\log\frac{2p(i|x)}{q_y(i) + p(i|x)} \nonumber\\
&\quad\quad \times \left[\nabla_x f_i(x) - \sum_{j=1}^N p(j|x)\nabla_x f_j(x)\right].
\end{aligned}
\label{eq:js_gradient}
\end{equation}
where $g_i(x) = \nabla_x f_i(x) - \sum_j p(j|x)\nabla_x f_j(x)$.
\end{corollary}

The symmetric nature of JS divergence provides several advantages. Unlike forward KL (which heavily penalizes $p$ for spreading beyond $q_y$) and reverse KL (which heavily penalizes $p$ for missing modes in $q_y$), JS divergence applies moderate penalties in both directions through the mixture $m$. The weighting term $(q_y(i) - p(i|x))/m(i)$ is bounded and becomes small when $q_y(i) \approx p(i|x)$, providing smooth gradient dynamics. This may offer a favorable trade-off between maintaining target class fidelity and preserving sample diversity, as we verify empirically in Section~\ref{exp}.

The overall algorithm for $f$-divergence guided sampling is provided in Algorithm~\ref{alg:f-divergence-guided}.

\begin{algorithm}
\caption{DDPM $f$-Divergence Guided Sampling}
\label{alg:f-divergence-guided}
\begin{algorithmic}[1]
\Require Diffusion model $\mathcal{D}_\theta$, classifier $f$, class label $y$, temperatures $\tau_1, \tau_2$, divergence weight $\alpha$, target bias $\epsilon$, target distribution $q_y$, classifier guidance scale $\gamma_t$, divergence type $\mathcal{D}$.
\State $\hat{x}_T \sim \mathcal{N}(0, I)$
\For{$t = T, \ldots, 1$}
    \State $\mu, \epsilon_\theta(\hat{x}_t, y, t) \leftarrow \mathcal{D}_\theta(\hat{x}_t, y, t)$
    \State $\hat{x}_0(t) \leftarrow (\hat{x}_t - \sqrt{1-\bar{\alpha}_t} \epsilon_\theta(\hat{x}_t, y, t)) / \sqrt{\bar{\alpha}_t}$ \Comment{Predicted clean sample}
    \State $p \leftarrow \text{softmax}(f(\hat{x}_0(t)))$ \Comment{Current distribution}
    \State Compute $g \leftarrow \nabla_{\hat{x}_0(t)} \mathcal{S}_{\mathcal{D}}(x, y)$ 
    \State $\hat{x}_{t-1} \sim \mathcal{N}(\mu + \gamma_t g, \sigma_t)$
\EndFor
\State \Return $\hat{x}_0$
\end{algorithmic}
\end{algorithm}

\subsection{Tilted sampling}

Since we are using batch sampling, we can leverage the information across the samples generated in each batch. This batch-aware sampling could lead to better adjustments for outlier-generated images. However, we do not want to add extra complexity to the sampling process. We utilize the loss introduced in \cite{li2023tiltedlossesmachinelearning} for coefficient adjustment in our sampling process:
\begin{equation}
\mathcal{S}_{tilted}(t; x,y) := \frac{1}{t} \log \left( \frac{1}{N} \sum_{i \in [N]} e^{t \: \log p_{\tau_1,\tau_2}(y|x)} \right).
\end{equation}
where $y$ is the target class we want to generate. 
Following \cite{li2023tiltedlossesmachinelearning}, the derivative will have a coefficient which is a distribution over samples.
The new derivative  is $g^i_{new} =w^i(t; \theta)  \left. \nabla_{x_t} \log p_{\tau_1,\tau_2}(y|x)\right|_{x_t=\mu}$ (details in the Appendix \ref{tilted}). This formulation allows us to control the diversity of generated samples by tuning the parameter $t$. When $t$ is large and positive, the process gives more weight to high-probability samples and pays less attention to outliers, which improves sample quality but reduces diversity. In contrast, when $t$ is negative, the process emphasizes low-probability or outlier samples, which increases diversity but may reduce overall sample quality. In this way, the hyperparameter $t$ provides a direct trade-off between quality and diversity.

\section{Experiments} \label{exp}

\textbf{Smooth ECE}: We evaluate the effectiveness of the Smooth ECE regularizer (for details see Appendix~\ref{eceimp}) when applied during fine-tuning. As shown in Table~\ref{tab:ece}, incorporating the Smooth ECE regularization leads to a reduction of 0.2 in the FID score, corresponding to a relative improvement of approximately 3\%. These results show that the Smooth ECE regularizer not only improves the quality of generated samples but could also be useful as a regularization method during the main training phase (to be investigated in the future work). Moreover, the observed increase in precision indicates a positive shift towards more class-consistent samples, while recall remains competitive, showing that the regularization does not excessively constrain diversity.

\begin{table}[h]
  \centering
  \vspace{-1em}
  \caption{Comparison of the baseline DDPM, classifier-guided diffusion, and our Smooth ECE-guided diffusion. All models are sampled for 250 DDPM steps, and we generate 10k ImageNet $128\times128$ samples for evaluation. Lower is better for FID; higher is better for Precision and Recall. FID values are averaged over three runs.}
  \vspace{5pt}
  \label{tab:ece}
  \begin{tabular}{lcccc}
    \toprule
    \textbf{Method} & \textbf{Classifier} & \textbf{FID} $\downarrow$ & \textbf{Precision} $\uparrow$ & \textbf{Recall} $\uparrow$ \\
    \midrule
    \cite{dhariwal2021diffusion}   & Basic fine-tuned          & $6.15$ & $0.77$ & $\mathbf{0.68}$ \\
    Diffusion + ECE (ours)         & ECE fine-tuned            & $\mathbf{5.94}$ & $\mathbf{0.79}$ & $0.66$ \\
    \bottomrule
  \end{tabular}
\end{table}

\textbf{Sampling Guidance}: We explore three different sampling strategies to improve inference:  (1) \textit{adaptive entropy-regularized sampling}; and (2) \textit{divergence-regularized sampling}; (3) \textit{tilted sampling}, which incorporates batch-level information using a tilted loss. As shown in Table~\ref{tab:fid10koff}, all three approaches outperform the baseline guided by ResNet-50~\cite{ma2023elucidating}. Tilted sampling (with $t = -0.2$) slightly reduces the FID from 5.335 to 5.281, demonstrating the value of leveraging batch-level guidance. Entropy-regularized sampling also leads to modest improvements in both FID and recall. As we anticipated, this method has the best recall value by incorporating the class-agnostic regularization. The best performance comes from divergence-guided sampling, which achieves the lowest FID score of 5.118. The divergence-regularized method attains the lowest FID while maintaining strong recall and precision, indicating that balancing class consistency with exploration enhances both quality and coverage.
\begin{table}[h]
\vspace{-1em}
  \centering
  \caption{The comparison of the baseline DDPM diffusion and our regularized guided sampling. All models are sampled for 250 DDPM steps. We generate 10k ImageNet $128\times128$ samples for evaluation. FID values are averaged over three runs.}
\vspace{5pt}
  \label{tab:fid10koff}
  \begin{tabular}{lcccc}
    \toprule
    \textbf{Method} & \textbf{Classifier} &\textbf{FID} $\downarrow$ & \textbf{Precision} $\uparrow$ & \textbf{Recall} $\uparrow$ \\
    \midrule
    \cite{ma2023elucidating} & ResNet-50 &  $\,5.34 $ & $\mathbf{0.78} $ & $0.67 $ \\
    Diffusion tilted guided ($t=-0.2$) (ours) & ResNet-50  &  $\,5.28 $ & $0.77 $ & $0.68 $ \\
    Diffusion adaptive entropy guided (ours) & ResNet-50  &  $\,5.30 $ & $0.77 $ & $\mathbf{0.69} $ \\
    RKL guided (ours) & ResNet-50  &  $\,\mathbf{5.12}$ & $\mathbf{0.78} $ & $0.68 $ \\
    \bottomrule
  \end{tabular}
\end{table}

Lastly, Table~\ref{tab:fid50koff} benchmarks our best-performing approaches against representative state-of-the-art methods on ImageNet $128{\times}128$ with the matched sampling steps, without retraining.
\begin{table}[h]
	\centering
    \vspace{-1em}
	\caption{The comparison of the baseline DDPM diffusion and our divergence guided samplings. All models are sampled for 250 DDPM steps. We generate 50k ImageNet $128\times128$ samples for evaluation.}
	\vspace{5pt}
	\label{tab:fid50koff}
	\begin{tabular}{lcccc}
		\toprule
		\textbf{Method} & \textbf{Classifier} &\textbf{FID} $\downarrow$ & \textbf{Precision} $\uparrow$ & \textbf{Recall} $\uparrow$ \\
		\midrule
		\cite{dhariwal2021diffusion} & Basic fine-tuned &  $\,2.97 $ & $\,0.78$&$\,0.59$\\
        \cite{zheng2022entropy} & Entropy aware classifier &  $\,2.68 $ & $\,0.80$&$\,0.56$\\
		\cite{ho2022classifier} & - &  $\,2.43 $ & -&-\\
		\cite{ma2023elucidating} & ResNet-50 &  $\,2.37 $ &$\,0.77$ &$\,0.60$\\
		FKL guided (ours) & ResNet-50  &  $\,2.33$ & $\,0.78$&$\,0.61$\\
        RKL guided (ours) & ResNet-50  &  $\,2.29$ & $\,0.78$&$\,0.61$\\
        JS guided (ours) & ResNet-50  &  $\,2.27$ & $\,0.77$&$\,\mathbf{0.62}$\\
        \cite{ma2023elucidating} & ResNet-101 &  $\,2.19 $ &$\,0.79 $ &$\,0.58 $\\
		FKL guided (ours) & ResNet-101  &  $\,2.17$ &$\,\mathbf{0.80} $ &$\,0.59 $\\
        RKL guided (ours) & ResNet-101  &  $\,2.14$ &$\,0.79 $ &$\,0.59 $\\
        JS guided (ours) & ResNet-101  &  $\,\mathbf{2.13}$ &$\,0.79 $ &$\,0.60 $\\
		\bottomrule
	\end{tabular}
\end{table}
All three $f$-divergence methods substantially outperform the baseline \citep{ma2023elucidating}, with Jensen-Shannon divergence achieving the best performance (FID of 2.27 with ResNet-50 and \textbf{2.13 with ResNet-101}), establishing a new result without any need to retrain the classifier or diffusion model.

The consistent ordering JS $>$ RKL $>$ FKL $>$ Baseline can be explained by examining how each divergence balances precision and recall. Forward KL's mode-seeking behavior strongly penalizes $p(\cdot|x)$ for placing mass outside $q_y$'s support, forcing the model to concentrate on high-confidence regions. While this yields the highest precision (0.80 with ResNet-101), it comes at the cost of the lowest recall (0.59), confirming that aggressive mode-seeking sacrifices diversity. The logarithmic weighting term $\log(q_y(i)/p(i|x))$ in Equation~\eqref{eq:fkl_gradient} creates strong repulsive forces that prematurely collapse the sampling distribution.

Reverse KL exhibits the opposite behavior: its mode-covering property (Lemma~\ref{lem:rkl_decomposition}) maintains broader support, yielding better recall (0.59) while slightly sacrificing precision (0.79). However, the linear weighting $q_y(i)$ in Equation~\eqref{eq:rkl_gradient} can lead to overshooting when $q_y$ is diffuse, as the regularization pulls toward \emph{all} classes weighted by $q_y$ rather than adaptively moderating based on current prediction quality.

Jensen-Shannon achieves the best balance through its mixture distribution $m(i) = \frac{1}{2}(q_y(i) + p(i|x))$. The symmetric weighting $(q_y(i) - p(i|x))/m(i)$ in Equation~\eqref{eq:js_gradient} provides \emph{adaptive} corrections that strengthen when $p$ deviates from $q_y$ but moderate when they align. This dynamic penalty structure achieves the highest recall (0.60) while maintaining competitive precision (0.79), confirming that JS's balanced approach to mode-seeking and mode-covering is optimal for diffusion guidance. The mixture $m$ acts as a stabilizing reference that prevents both premature collapse (FKL's failure mode) and excessive exploration (RKL's tendency).
ResNet-101 also improves our best result from 2.27 to 2.13 FID, while maintaining consistent relative improvements over the baseline (4-8\% FID reduction).

\section{Conclusion}

This work addresses critical limitations in classifier-guided diffusion through calibration-aware design and regularized sampling guidance. Our key contributions include a differentiable Smooth ECE loss achieving 3\% FID improvement with minimal overhead, three sampling methods with divergence guidance achieving the best FID of 2.13, and theoretical analysis providing mathematical grounding for observed behaviors.
This work demonstrates that principled $f$-divergence regularization can substantially improve classifier-guided diffusion without model retraining, achieving FID of 2.13 on ImageNet 128$\times$128. Our theoretical framework reveals how different divergence choices (RKL, FKL, JS) trade off mode coverage and precision, with Jensen-Shannon's balanced penalties proving empirically optimal. These contributions offer practical, plug-and-play enhancements for conditional generation in deployed systems where diffusion models cannot be retrained with more complex approaches.
\bibliographystyle{plainnat}
\bibliography{refs}

@article{
bohdal2023metacalibration,
title={Meta-Calibration: Learning of Model Calibration Using Differentiable Expected Calibration Error},
author={Ondrej Bohdal and Yongxin Yang and Timothy Hospedales},
journal={Transactions on Machine Learning Research},
issn={2835-8856},
year={2023},
url={https://openreview.net/forum?id=R2hUure38l},
note={}
}

@inproceedings{zheng2022entropy,
  title={Entropy-driven sampling and training scheme for conditional diffusion generation},
  author={Zheng, Guangcong and Li, Shengming and Wang, Hui and Yao, Taiping and Chen, Yang and Ding, Shouhong and Li, Xi},
  booktitle={European Conference on Computer Vision},
  pages={754--769},
  year={2022},
  organization={Springer}
}

@article{calafiore2019log,
  title={Log-sum-exp neural networks and posynomial models for convex and log-log-convex data},
  author={Calafiore, Giuseppe C and Gaubert, Stephane and Possieri, Corrado},
  journal={IEEE transactions on neural networks and learning systems},
  volume={31},
  number={3},
  pages={827--838},
  year={2019},
  publisher={IEEE}
}

@book{murphy2012machine,
  title={Machine learning: a probabilistic perspective},
  author={Murphy, Kevin P},
  year={2012},
  publisher={MIT press}
}

@article{williams1998bayesian,
  title={Bayesian classification with Gaussian processes},
  author={Williams, Christopher KI and Barber, David},
  journal={IEEE Transactions on pattern analysis and machine intelligence},
  volume={20},
  number={12},
  pages={1342--1351},
  year={1998},
  publisher={IEEE}
}

@inproceedings{behnamnialog,
  title={Log-Sum-Exponential Estimator for Off-Policy Evaluation and Learning},
  author={Behnamnia, Armin and Aminian, Gholamali and Aghaei, Alireza and Shi, Chengchun and Tan, Vincent YF and Rabiee, Hamid R},
  booktitle={Forty-second International Conference on Machine Learning}
}

@article{li2023tiltedlossesmachinelearning,
  author  = {Tian Li and Ahmad Beirami and Maziar Sanjabi and Virginia Smith},
  title   = {On Tilted Losses in Machine Learning: Theory and Applications},
  journal = {Journal of Machine Learning Research},
  year    = {2023},
  volume  = {24},
  number  = {142},
  pages   = {1--79},
  url     = {http://jmlr.org/papers/v24/21-1095.html}
}

@article{song2020score,
  title={Score-based generative modeling through stochastic differential equations},
  author={Song, Yang and Sohl-Dickstein, Jascha and Kingma, Diederik P and Kumar, Abhishek and Ermon, Stefano and Poole, Ben},
  journal={arXiv preprint arXiv:2011.13456},
  year={2020}
}

@article{ho2020denoising,
  title={Denoising diffusion probabilistic models},
  author={Ho, Jonathan and Jain, Ajay and Abbeel, Pieter},
  journal={Advances in neural information processing systems},
  volume={33},
  pages={6840--6851},
  year={2020}
}

@inproceedings{nichol2021improved,
  title={Improved denoising diffusion probabilistic models},
  author={Nichol, Alexander Quinn and Dhariwal, Prafulla},
  booktitle={International conference on machine learning},
  pages={8162--8171},
  year={2021},
  organization={PMLR}
}

@article{song2020denoising,
  title={Denoising diffusion implicit models},
  author={Song, Jiaming and Meng, Chenlin and Ermon, Stefano},
  journal={arXiv preprint arXiv:2010.02502},
  year={2020}
}

@article{dhariwal2021diffusion,
  title={Diffusion models beat gans on image synthesis},
  author={Dhariwal, Prafulla and Nichol, Alexander},
  journal={Advances in neural information processing systems},
  volume={34},
  pages={8780--8794},
  year={2021}
}

@inproceedings{sohl2015deep,
  title={Deep unsupervised learning using nonequilibrium thermodynamics},
  author={Sohl-Dickstein, Jascha and Weiss, Eric and Maheswaranathan, Niru and Ganguli, Surya},
  booktitle={International conference on machine learning},
  pages={2256--2265},
  year={2015},
  organization={pmlr}
}

@article{karras2022elucidating,
  title={Elucidating the design space of diffusion-based generative models},
  author={Karras, Tero and Aittala, Miika and Aila, Timo and Laine, Samuli},
  journal={Advances in neural information processing systems},
  volume={35},
  pages={26565--26577},
  year={2022}
}

@article{csiszar1967information,
  title={On information-type measure of difference of probability distributions and indirect observations},
  author={Csisz{\'a}r, Imre},
  journal={Studia Sci. Math. Hungar.},
  volume={2},
  pages={299--318},
  year={1967}
}

@article{ma2023elucidating,
  title={Elucidating the design space of classifier-guided diffusion generation},
  author={Ma, Jiajun and Hu, Tianyang and Wang, Wenjia and Sun, Jiacheng},
  journal={arXiv preprint arXiv:2310.11311},
  year={2023}
}

@book{polyanskiy2025information,
  title={Information theory: From coding to learning},
  author={Polyanskiy, Yury and Wu, Yihong},
  year={2025},
  publisher={Cambridge university press}
}

@inproceedings{ramesh2021zero,
  title={Zero-shot text-to-image generation},
  author={Ramesh, Aditya and Pavlov, Mikhail and Goh, Gabriel and Gray, Scott and Voss, Chelsea and Radford, Alec and Chen, Mark and Sutskever, Ilya},
  booktitle={International conference on machine learning},
  pages={8821--8831},
  year={2021},
  organization={Pmlr}
}

@inproceedings{rombach2022high,
  title={High-resolution image synthesis with latent diffusion models},
  author={Rombach, Robin and Blattmann, Andreas and Lorenz, Dominik and Esser, Patrick and Ommer, Bj{\"o}rn},
  booktitle={Proceedings of the IEEE/CVF conference on computer vision and pattern recognition},
  pages={10684--10695},
  year={2022}
}

@article{liu2024sora,
  title={Sora: A review on background, technology, limitations, and opportunities of large vision models},
  author={Liu, Yixin and Zhang, Kai and Li, Yuan and Yan, Zhiling and Gao, Chujie and Chen, Ruoxi and Yuan, Zhengqing and Huang, Yue and Sun, Hanchi and Gao, Jianfeng and others},
  journal={arXiv preprint arXiv:2402.17177},
  year={2024}
}

@inproceedings{song2023consistency,
  title={Consistency Models},
  author={Song, Yang and Dhariwal, Prafulla and Chen, Mark and Sutskever, Ilya},
  booktitle={Proceedings of the 40th International Conference on Machine Learning (ICML)},
  year={2023},
  volume={202},
  pages={31999--32015}
}

@inproceedings{song2019generative,
  title={Generative Modeling by Estimating Gradients of the Data Distribution},
  author={Song, Yang and Ermon, Stefano},
  booktitle={Advances in Neural Information Processing Systems (NeurIPS)},
  year={2019},
  volume={32}
}

@inproceedings{naeini2015obtaining,
  title={Obtaining well calibrated probabilities using bayesian binning},
  author={Naeini, Mahdi Pakdaman and Cooper, Gregory and Hauskrecht, Milos},
  booktitle={Proceedings of the AAAI conference on artificial intelligence},
  volume={29},
  number={1},
  year={2015}
}

@article{heusel2017gans,
  title={Gans trained by a two time-scale update rule converge to a local nash equilibrium},
  author={Heusel, Martin and Ramsauer, Hubert and Unterthiner, Thomas and Nessler, Bernhard and Hochreiter, Sepp},
  journal={Advances in neural information processing systems},
  volume={30},
  year={2017}
}

@book{bishop2006pattern,
  title={Pattern recognition and machine learning},
  author={Bishop, Christopher M and Nasrabadi, Nasser M},
  volume={4},
  number={4},
  year={2006},
  publisher={Springer}
}

@article{ho2022classifier,
  title={Classifier-free diffusion guidance},
  author={Ho, Jonathan and Salimans, Tim},
  journal={arXiv preprint arXiv:2207.12598},
  year={2022}
}

@article{kawar2022enhancing,
  title={Enhancing diffusion-based image synthesis with robust classifier guidance},
  author={Kawar, Bahjat and Ganz, Roy and Elad, Michael},
  journal={arXiv preprint arXiv:2208.08664},
  year={2022}
}

@inproceedings{peebles2023scalable,
  title={Scalable diffusion models with transformers},
  author={Peebles, William and Xie, Saining},
  booktitle={Proceedings of the IEEE/CVF international conference on computer vision},
  pages={4195--4205},
  year={2023}
}

@article{li2020tilted,
  title={Tilted empirical risk minimization},
  author={Li, Tian and Beirami, Ahmad and Sanjabi, Maziar and Smith, Virginia},
  journal={arXiv preprint arXiv:2007.01162},
  year={2020}
}

@incollection{huber2011robust,
  title={Robust statistics},
  author={Huber, Peter J},
  booktitle={International encyclopedia of statistical science},
  pages={1248--1251},
  year={2011},
  publisher={Springer}
}

@article{kingma2023understanding,
  title={Understanding diffusion objectives as the elbo with simple data augmentation},
  author={Kingma, Diederik and Gao, Ruiqi},
  journal={Advances in Neural Information Processing Systems},
  volume={36},
  pages={65484--65516},
  year={2023}
}

@article{xu2024disco,
  title={Disco-diff: Enhancing continuous diffusion models with discrete latents},
  author={Xu, Yilun and Corso, Gabriele and Jaakkola, Tommi and Vahdat, Arash and Kreis, Karsten},
  journal={arXiv preprint arXiv:2407.03300},
  year={2024}
}

@article{ali1966general,
  title={A general class of coefficients of divergence of one distribution from another},
  author={Ali, S Mussafer and Silvey, Samuel D},
  journal={Journal of the Royal Statistical Society: Series B (Methodological)},
  volume={28},
  number={1},
  pages={131--142},
  year={1966}
}

@article{sason2016divergence,
  title={f-divergence inequalities},
  author={Sason, Igal and Verd{\'u}, Sergio},
  journal={IEEE Transactions on Information Theory},
  volume={62},
  number={11},
  pages={5973--6006},
  year={2016}
}

@inproceedings{nowozin2016f,
  title={f-gan: Training generative neural samplers using variational divergence minimization},
  author={Nowozin, Sebastian and Cseke, Botond and Tomioka, Ryota},
  booktitle={Advances in Neural Information Processing Systems},
  volume={29},
  year={2016}
}

@inproceedings{arjovsky2017wasserstein,
  title={Wasserstein generative adversarial networks},
  author={Arjovsky, Martin and Chintala, Soumith and Bottou, L{\'e}on},
  booktitle={International Conference on Machine Learning},
  pages={214--223},
  year={2017}
}

@article{jordan1999introduction,
  title={An introduction to variational methods for graphical models},
  author={Jordan, Michael I and Ghahramani, Zoubin and Jaakkola, Tommi S and Saul, Lawrence K},
  journal={Machine Learning},
  volume={37},
  number={2},
  pages={183--233},
  year={1999}
}

@inproceedings{li2016renyi,
  title={R{\'e}nyi divergence variational inference},
  author={Li, Yingzhen and Turner, Richard E},
  booktitle={Advances in Neural Information Processing Systems},
  volume={29},
  year={2016}
}

@article{dieng2017chi,
  title={Variational inference via $\chi$ upper bound minimization},
  author={Dieng, Adji Bousso and Tran, Dustin and Ranganath, Rajesh and Paisley, John and Blei, David},
  journal={Advances in Neural Information Processing Systems},
  volume={30},
  year={2017}
}

@inproceedings{song2021maximum,
  title={Maximum likelihood training of score-based diffusion models},
  author={Song, Yang and Durkan, Conor and Murray, Iain and Ermon, Stefano},
  booktitle={Advances in Neural Information Processing Systems},
  volume={34},
  pages={1415--1428},
  year={2021}
}

@inproceedings{nachum2017trust,
  title={Trust-pcl: An off-policy trust region method for continuous control},
  author={Nachum, Ofir and Norouzi, Mohammad and Xu, Kelvin and Schuurmans, Dale},
  booktitle={International Conference on Learning Representations},
  year={2017}
}

@inproceedings{lee2021optidice,
  title={Optidice: Offline policy optimization via stationary distribution correction estimation},
  author={Lee, Jongmin and Jeon, Wonseok and Lee, Byung-Jun and Pineau, Joelle and Kim, Kee-Eung},
  booktitle={International Conference on Machine Learning},
  pages={6120--6130},
  year={2021}
}


\appendix

\section{Related Works}
\textbf{Other improved guidance approaches}: \cite{zheng2022entropy} study the vanishing-guidance problem in classifier-guided DDPMs and propose an entropy-based scheme that weights the classifier gradient by a function of predictive entropy during sampling. This promotes exploration and helps preserve structural fidelity, yielding empirical gains in conditional generation. However, the approach leaves the guidance direction unchanged and is not supported by a general theoretical analysis for arbitrary classifiers. As we discussed in Section~\ref{rsample}, the absence of theoretical analysis and the limited effectiveness of this method for general classifiers motivated our work, which directly regularizes the sampling objective. \cite{kawar2022enhancing} proposed to harness the perceptually aligned gradients phenomenon by utilizing robust classifiers to guide a diffusion process, while this method showed a minor improvement compared to \cite{dhariwal2021diffusion}, the limited contribution could not surpass our second baseline \cite{ma2023elucidating}. 

 
\textbf{Classifier-free guidance}: Rather than relying on an external classifier, \cite{ho2022classifier} proposed classifier-free guidance in which they train a single denoiser to predict diffusion noise both \emph{with} a condition and \emph{without} it. The gap between these two predictions approximates the conditional score, which can be used to steer sampling. In practice, guidance is applied by 
$
\epsilon^{*}_{\theta}(x_t,y,t)
= \epsilon_{\theta}(x_t,y,t)
+ (s-1)\Big[\epsilon_{\theta}(x_t,y,t)-\epsilon_{\theta}(x_t,\varnothing,t)\Big],
$ 
where \(s\!\ge\!1\) is the guidance scale and \(\epsilon_{\theta}(x_t,\varnothing,t)\) denotes the unconditional prediction learned by randomly dropping the condition during training. This simple setup avoids an auxiliary classifier and typically improves on-condition fidelity in generated samples.

\textbf{Beyond classifier guidance}: The diffusion modeling landscape has seen significant advances across multiple fronts. Score-based generative models \cite{song2020score} provide an alternative mathematical framework using stochastic differential equations (SDEs), enabling continuous-time formulations and advanced sampling techniques like probability flow ordinary differential equation (ODE). Denoising diffusion implicit models (DDIM) \cite{song2020denoising} introduced deterministic sampling that dramatically reduces the number of denoising steps required. Latent diffusion models \cite{rombach2022high} operate in compressed latent spaces, significantly reducing computational costs while maintaining generation quality. EDM \cite{karras2022elucidating} unified various diffusion formulations and introduced improved training and sampling strategies. Consistency models \cite{song2023consistency} enable single-step generation through distillation from pre-trained diffusion models. DiT \cite{peebles2023scalable} replaced U-Net architectures with transformers, achieving state-of-the-art results on class-conditional ImageNet generation. More recent diffusion approaches have reported improved FID scores by reformulating diffusion objectives or augmenting the latent space. \citet{kingma2023understanding} reinterpret common diffusion losses as ELBO variants with monotonic weighting, achieving state-of-the-art FID on ImageNet. \citet{xu2024disco} propose DisCo-Diff, which introduces discrete latents to simplify the noise-to-data mapping, further improving image quality. While effective, both methods require retraining diffusion models from scratch and often rely on longer sampling chains (nearly doubling the diffusion steps), making them more computational demanding in practice.

\textbf{Tilted losses and robust optimization}: Inspired by the log-sum-exponential operator with applications in multinomial linear regression, naive Bayes classifiers, tilted empirical risk, and log-sum-exponential off-policy estimator,~\citep{calafiore2019log,murphy2012machine,williams1998bayesian,li2020tilted,behnamnialog}, we propose the tilted sampling. The tilted loss framework introduced by \cite{li2020tilted, li2023tiltedlossesmachinelearning} provides a principled approach for controlling the influence of outliers through a temperature parameter t, enabling weighted gradient computation where sample contributions are modulated by their relative importance. This framework extends beyond simple outlier control, offering connections to distributionally robust optimization (DRO) and risk-sensitive learning. In the context of machine learning, tilted losses have been successfully applied to federated learning for handling heterogeneous data distributions, reinforcement learning for risk-aware policy optimization, and meta-learning for improved generalization across tasks. The temperature parameter t in tilted losses creates a spectrum between average-case ($t \to 0$) and worst-case ($t \to \infty$) optimization, with negative values emphasizing easy examples—a property we exploit in our diffusion guidance to down-weight overconfident classifier predictions.

\textbf{$f$-divergences}: $f$-divergences \cite{csiszar1967information, ali1966general} provide a unified framework for comparing probability distributions, encompassing KL divergence, Hellinger distance, and total variation as special cases \cite{sason2016divergence}. In generative modeling, $f$-GANs \cite{nowozin2016f} showed that different GAN formulations minimize different $f$-divergences, with forward KL causing mode-dropping and reverse KL encouraging mode-covering \cite{arjovsky2017wasserstein}—insights directly motivating our guidance design.

Variational inference has explored $f$-divergence alternatives beyond forward KL \cite{jordan1999introduction}, including Rényi \cite{li2016renyi} and $\chi$ divergences \cite{dieng2017chi}, demonstrating that divergence choice affects underfitting-overfitting trade-offs. Similar trade-offs appear in RL policy optimization \cite{nachum2017trust, lee2021optidice}, where divergence selection impacts exploration-exploitation balance.

In diffusion models, prior work analyzed $f$-divergences for \emph{training objectives} \cite{song2021maximum, kingma2023understanding}. Our work is the first systematic exploration of $f$-divergences for \emph{classifier-guided sampling}, where divergence acts as a regularizer rather than training objective. We provide: (1) rigorous gradient derivations (Proposition~\ref{prop:general_f_div}), (2) closed-form Gaussian analysis (Proposition~\ref{prop:gaussian_rkl}), and (3) empirical demonstration that Jensen-Shannon divergence achieves superior precision-recall balance. The finding that symmetric JS outperforms both forward and reverse KL challenges conventional wisdom that mode-covering (reverse KL) is universally optimal for generation, suggesting balanced penalties are preferable when diversity and fidelity are equally valued.


\section{Background}
In this section, we provide a brief overview of the formulation of Gaussian diffusion models as discussed in \cite{ho2020denoising} and the concept of classifier guidance referenced in \cite{dhariwal2021diffusion}.
\subsection{Gaussian diffusion models}
Diffusion models consist of a series of time-dependent components that implement both forward and reverse processes. We define data distribution as $x_0 \sim q(x_0)$ and a Markovian noising process $q$ which gradually adds noise to the data to produce noised samples $\{x_t\}_{t=1}^T$. This process is defined as the forward process, and for a given forward variance $\beta_t$, is defined as
\begin{equation}
    q\left(x_t \mid x_{t-1}\right):=\mathcal{N}\left(x_t ; \sqrt{1-\beta_t} x_{t-1}, \beta_t \mathbf{I}\right). 
\end{equation}
Using $\alpha_t:=1-\beta_t$ and $\bar{\alpha}_t:=\prod_{s=0}^t \alpha_s$ this can also be written as
\begin{equation}
    q\left(x_t \mid x_0\right)=\mathcal{N}\left(x_t ; \bar{\alpha}_t x_0,\left(1-\bar{\alpha}_t\right)\right).
\end{equation}
The Bayes' theorem will show that the posterior $q\left(x_{t-1} \mid x_t, x_0\right) $ has a Gaussian distribution with the following mean and variance: 
\begin{equation}
\tilde{\mu}_t\left(x_t, x_0\right)  :=\frac{\sqrt{\bar{\alpha}_{t-1}} \beta_t}{1-\bar{\alpha}_t} x_0+\frac{\sqrt{\alpha_t}\left(1-\bar{\alpha}_{t-1}\right)}{1- \bar{\alpha}_t} x_t,
\end{equation}
\begin{equation}
\tilde{\beta}_t  :=\frac{1-\bar{\alpha}_{t-1}}{1-\bar{\alpha}_t} \beta_t,
\end{equation}
\begin{equation}
    q\left(x_{t-1} \mid x_t, x_0\right)  =\mathcal{N}\left(x_{t-1} ; \tilde{\mu}\left(x_t, x_0\right), \tilde{\beta}_t \mathbf{I}\right).
\end{equation}
In the reverse process, clean samples are gradually generated from noisy samples. This process is defined as
\begin{equation}
    p_\theta\left(\widehat{x}_{t-1} \mid \widehat{x}_t\right)=\mathcal{N}\left(\widehat{x}_{t-1} ; \mu_\theta\left(\widehat{x}_t, t\right), \sigma_t\right),
\end{equation}
where $\mu_\theta\left(\widehat{x}_t, t\right)$ is derived from removing the diffusion estimated $\epsilon_\theta\left(\widehat{x}_t, t\right)$ from the noisy samples $\widehat{x}_t: \mu_\theta\left(\widehat{x}_t, t\right)=\frac{1}{\sqrt{\alpha_t}}\left(\widehat{x}_t-\frac{\beta_t}{\sqrt{1-\bar{\alpha}_t}} \epsilon_\theta\left(\widehat{x}_t, t\right)\right)$, and  $\sigma_t$ denotes the reverse process variance. To train this model so that $p(x_0)$ approximates the true data distribution $q(x_0)$, the following variational lower bound ($L_{\mathrm{vlb}}$), is optimized
\begin{equation}
L_{\mathrm{vlb}}  := \sum_{t=0}^T L_t ,
\end{equation}
\begin{equation}
L_0  :=-\log p_\theta\left(x_0 \mid x_1\right) ,
\end{equation}
\begin{equation}
L_{t-1} :=D_{K L}\left(q\left(x_{t-1} \mid x_t, x_0\right) \| p_\theta\left(x_{t-1} \mid x_t\right)\right),
\end{equation}
\begin{equation}
L_T :=D_{K L}\left(q\left(x_T \mid x_0\right) \| p\left(x_T\right)\right),
\end{equation}
where $D_{K L}(p(x)\| q(x))=\int_{\mathcal{X}} p(x)\log(p(x)/q(x))\mathrm{d}x$ is the KL-divergence between two distribution $p(x)$ and $q(x)$.
In practice, \cite{ho2020denoising} proposed a simplified training objective that
predicts the Gaussian noise added at step $t$ as
\begin{equation}
L_{\text{simple}}
:= \mathbb{E}_{t \sim \left[ 1,T\right],\, x_0 \sim q(x_0),\, \epsilon \sim \mathcal{N}(0,\mathbf{I})}
\big[\,\|\epsilon - \epsilon_\theta(x_t,t)\|_2^2\,\big].
\end{equation}
This parameterization trains the model as a noise predictor $\epsilon_\theta(x_t,t)$ rather than directly regressing the mean $\mu_\theta(\hat{x}_t,t)$, which is equivalent up to a constant factor in the variational objective and is empirically easier to optimize.

\subsection{Classifier guidance diffusion models}

Classifier guidance \cite{dhariwal2021diffusion} can be applied in the reverse process. This can guide the sampling trajectory toward regions of higher classifier likelihood and improve fidelity and class-consistency. We start with a diffusion model with an unconditional reverse noising process $ p_\theta\left(x_{t-1} \mid x_t\right)$. We
condition this on a label $y$, according to 
\begin{equation}
    p_{\theta, \phi}\left(x_t \mid x_{t+1}, y\right) \propto  p_\theta\left(x_t \mid x_{t+1}\right) p_\phi\left(y \mid x_t\right).
\end{equation}
$\log p_\phi\left(y \mid x_t\right)$ can be approximated using a Taylor expansion around $x_t = \mu$ as
\begin{equation}
    \begin{aligned}
\log p_\phi\left(y \mid x_t\right) & \left.\approx \log p_\phi\left(y \mid x_t\right)\right|_{x_t=\mu}+\left.\left(x_t-\mu\right) \nabla_{x_t} \log p_\phi\left(y \mid x_t\right)\right|_{x_t=\mu} \\
& =\left(x_t-\mu\right) g+C_1.
\end{aligned}
\end{equation}
Here, $g = \left. \nabla_{x_t} \log p_\phi\left(y \mid x_t\right)\right|_{x_t=\mu}$, and $C_1$ is a constant. This gives 
\begin{equation}
\label{noise_term}
    \begin{aligned}
\log \left(p_\theta\left(x_t \mid x_{t+1}\right) p_\phi\left(y \mid x_t\right)\right) & \approx-\frac{1}{2}\left(x_t-\mu\right)^T \Sigma^{-1}\left(x_t-\mu\right)+\left(x_t-\mu\right) g+C_2 \\
& =\log p(z)+C_3, z \sim \mathcal{N}(\mu+\Sigma g, \Sigma).
\end{aligned}
\end{equation}
It is thus established that the conditional transition operator can be approximated by a Gaussian distribution analogous to its unconditional counterpart, with the mean shifted by $\Sigma g$, where $g$ denotes the classifier gradient. In practice, this gradient is computed as $g = \nabla_{\widehat{x}_t} \log \left(p\left(y \mid \widehat{x}_t\right)\right)$, which we obtain by taking the gradient of the log-probability with respect to the noisy input. Specifically, for a classifier $f$ that outputs logits, we compute this as the gradient of the log-softmax $\nabla_{\widehat{x}_t} \log \left(\operatorname{softmax}\left(f_y\left(\widehat{x}_t\right)\right)\right)$, where $f_y\left(\widehat{x}_t\right)$ denotes the logit corresponding to class $y$.

\section{Supplementary Notes}

\subsection{Smooth ECE and Huber Loss} \label{ECE-Smooth}

The standard ECE measures the similarity  between predicted confidences and empirical accuracies, typically using an absolute error term:
\begin{equation}
    \mathcal{L}_{\mathrm{ECE}}^{(i)} = \left| \hat{p}^{(i)} - a^{(i)} \right|,
\end{equation}
where $\hat{p}^{(i)} = \max_{c} p_\phi(c \mid x^{(i)})$ denotes the model's predicted confidence, and $a^{(i)}$ indicates whether the prediction was correct.

This formulation, while simple and interpretable, is not differentiable at zero and may lead to instability or optimization challenges when used as a training loss.

To address this, we introduce a smoothed version:
\begin{equation}
    \mathcal{L}_{\mathrm{SmoothECE}}^{(i)} = \sqrt{ \left( \hat{p}^{(i)} - a^{(i)} \right)^2 + \beta },
\end{equation}
where $\beta > 0$ is a small constant that controls the degree of smoothing.

This formulation closely resembles the \textbf{Huber loss}, a well-known technique for robust regression that combines the benefits of both $\ell_1$ and $\ell_2$ losses \cite{huber2011robust}. The Huber loss is defined as
\begin{equation}
    \mathcal{L}_\delta(r) =
\begin{cases}
\frac{1}{2} r^2 & \text{if } |r| \leq \delta, \\
\delta \left( |r| - \frac{1}{2} \delta \right) & \text{otherwise},
\end{cases}
\end{equation}
where $r = \hat{p}^{(i)} - a^{(i)}$ is the calibration residual. The Huber loss behaves quadratically near zero (like $\ell_2$) and linearly in the tails (like $\ell_1$), which makes it robust to outliers while maintaining differentiability around the minimum.

\begin{itemize}
    \item \textbf{Quadratic behavior near zero:} When $|\hat{p}^{(i)} - a^{(i)}| \ll \sqrt{\beta}$, we can use a Taylor expansion:
\begin{equation}
    \sqrt{r^2 + \beta} \approx \sqrt{\beta} + \frac{1}{2\sqrt{\beta}} r^2 + \mathcal{O}(r^4),
\end{equation}
    which behaves like a scaled $\ell_2$ loss with a constant offset.
    
    \item \textbf{Linear behavior for large residuals:} When $|\hat{p}^{(i)} - a^{(i)}| \gg \sqrt{\beta}$, we get:
\begin{equation}
    \sqrt{r^2 + \beta} \approx |r| + \frac{\beta}{2 |r|} + \mathcal{O}\left(\frac{1}{|r|^3}\right),
\end{equation}
    which converges to $\ell_1$ behavior as $|r|$ increases.
\end{itemize}

Therefore, the Smooth ECE loss serves as a fully differentiable, Huber-like alternative to the standard ECE, with the added benefit of avoiding discontinuities in gradients.

\subsection{Tilted sampling \& Gradient weight} \label{tilted}
We want to use the information from all samples in a batch in a smart way.  
For this, we apply the \emph{tilted loss} introduced in \cite{li2023tiltedlossesmachinelearning}.  
This loss has a parameter $t$ which controls how much we focus on certain samples.  
By changing $t$, we can either put more weight on common (high-probability) samples, or on rare (outlier) samples.

The tilted loss is defined as
\begin{align}
\mathcal{L}(t; \theta) := \frac{1}{t} \log \left( \frac{1}{N} \sum_{i=1}^N e^{t \,\ell(f(x_i), y; \theta)} \right)
\end{align}
where $\ell(f(x_i), y; \theta)$ is any loss function.

In our case, we replace the loss with the log-likelihood score.  
This gives
\begin{equation}
\mathcal{S}_{tilted}(t;x,y) := \frac{1}{t} \log \left( \frac{1}{N} \sum_{i=1}^N e^{t \, \log p_{\tau_1,\tau_2}(y|x_i)} \right)
\end{equation}

\paragraph{Gradient form:}
Taking the gradient with respect to the  $x$ gives
\begin{align}
\nabla_{x} \mathcal{S}_{tilted}(t; x, y) 
= \frac{\sum_{i=1}^N e^{t  \log p_{\tau_1,\tau_2}(y|x_i)} \, \nabla_{x} \log p_{\tau_1,\tau_2}(y|x_i)}
        {\sum_{j=1}^N e^{t  \log p_{\tau_1,\tau_2}(y|x_j)}}
\end{align}

This can be written as a weighted sum of per-sample gradients:
\begin{align}
\nabla_{x} \mathcal{S}_{tilted}(t; x, y) 
= \sum_{i=1}^N w^i(t; x, y) \,\nabla_{x}  \log p_{\tau_1,\tau_2}(y|x_i)
\end{align}
with weights
\begin{align}
w^i(t; x, y) := \frac{e^{t  \log p_{\tau_1,\tau_2}(y|x_i)}}{\sum_{j=1}^N e^{t \log p_{\tau_1,\tau_2}(y|x_j)}}
\end{align}

The weights $w^i(t; x, y)$ form a normalized distribution over the batch. If $t>0$, high-probability samples get larger weights (mode-seeking). If $t<0$, low-probability samples get larger weights (outlier-seeking). If $t=0$, all samples have equal weight and follow the regular guidance.

\paragraph{Applying to classifier guidance:}
The tilted loss was first defined for parameter updates, but we can use the same idea for classifier gradients with respect to the sample $x$.  
This gives the tilted classifier gradient
\begin{align}
g^i_{\text{tilted}} 
= w^i(t; x, y)\,\left. \nabla_{x}  \log p_{\tau_1,\tau_2}(y|x_i)\right|_{x_t=\mu}
\end{align}

As we know from Eq. \ref{energy-eq1} we achieve 

\begin{align}
g^i_{\text{tilted}} 
= w^i(t; x, y)\,\left.[-\nabla_{x} E_{\tau_1}(x, y)+\nabla_{x} E_{\tau_2}(x) ] \right|_{x_t=\mu}
\end{align}
\paragraph{Sampling step:}
The modified sampling rule becomes
\begin{align}
z^i \sim \mathcal{N}\!\big(\mu + \Sigma g^i_{\text{tilted}}, \; \Sigma\big)
\end{align}

\medskip
In this way, the hyperparameter $t$ directly controls the trade-off between 
\emph{quality} (when $t>0$, we push towards high-likelihood samples) and 
\emph{diversity} (when $t<0$, we encourage exploring lower-likelihood samples).

\subsection{Introduction to $f$-divergences}
\label{app:fdiv}

$f$-divergences \citep{csiszar1967information, ali1966general} provide a general framework for measuring dissimilarity
between probability distributions. In this work we adopt the discrete convention
\begin{equation}
D_f(q \| p) \;=\; \sum_{i=1}^N q(i)\, f\!\left(\frac{p(i)}{q(i)}\right),
\label{eq:fdiv_discrete_convention}
\end{equation}
where $f:\mathbb{R}_+\to\mathbb{R}$ is convex and satisfies $f(1)=0$. Convexity implies $D_f(q\|p)\ge 0$ with equality
iff $p=q$.

Different choices of $f$ recover classical divergences; Table~\ref{tab:f_divergences} summarizes the divergences used.

\begin{table}[h]
\centering
\caption{$f$-divergences used in this work with their generator functions.}
\label{tab:f_divergences}
\begin{tabular}{llll}
\toprule
Divergence & $f(t)$ & $f'(t)$ & Notes \\
\midrule
Reverse KL ($D_{\mathrm{KL}}(q\|p)$) & $-\log t$ & $-1/t$ & penalizes $p\!\to\!0$ when $q\!>\!0$ \\
Forward KL ($D_{\mathrm{KL}}(p\|q)$) & $t\log t$ & $\log t + 1$ & penalizes $p\!>\!0$ when $q\!\to\!0$ \\
Jensen--Shannon & $-(t+1)\log\frac{t+1}{2} + t\log t$ & $\log\frac{2t}{t+1}$ & symmetric, bounded \\
Squared Hellinger & $(\sqrt{t}-1)^2$ & $1-\frac{1}{\sqrt t}$ & $H$ is a metric (not $H^2$) \\
\bottomrule
\end{tabular}
\end{table}

\paragraph{Gradient computation.}
Let $p(\cdot|x)$ denote the classifier output over $N$ classes and $q_y$ a fixed target distribution.
For \eqref{eq:fdiv_discrete_convention}, define
\begin{equation}
w_f(q,p) \;:=\; p\, f'\!\left(\frac{p}{q}\right).
\label{eq:wf_def}
\end{equation}
Using $\nabla_x p(i|x)=p(i|x)\nabla_x\log p(i|x)$, we obtain
\begin{equation}
\nabla_x D_f(q_y\|p(\cdot|x))
\;=\;
\sum_{i=1}^N w_f\big(q_y(i),p(i|x)\big)\,\nabla_x\log p(i|x),
\label{eq:fdiv_grad_general}
\end{equation}
and with the softmax identity
\(
\nabla_x\log p(i|x)=\nabla_x f_i(x)-\sum_j p(j|x)\nabla_x f_j(x),
\)
this yields Proposition~\ref{prop:general_f_div}.

\paragraph{Properties and intuition.}
Most $f$-divergences are asymmetric, while JS and Hellinger are symmetric.
The choice of $f$ determines the weighting $w_f(q_y(i),p(i|x))$ in \eqref{eq:fdiv_grad_general}:
for reverse KL, $w_f(q,p)=-q$ (non-vanishing as $p\to 0$);
for forward KL, $w_f(q,p)=p\big(\log\frac{p}{q}+1\big)$ (vanishes as $p\to 0$);
for squared Hellinger, $w_f(q,p)=p-\sqrt{pq}$ (damped by the geometric mean term).
These weightings help explain the different empirical precision--recall behaviors observed in Section~\ref{exp}.

\subsection{Squared Hellinger Divergence}
\label{app:hellinger}

For completeness, we also evaluated squared Hellinger distance, defined by
$f(t) = (\sqrt{t} - 1)^2$, with derivative $f'(t)=1-\frac{1}{\sqrt{t}}$.
Under our convention $t=\frac{p}{q}$, the corresponding weight is
$w_f(q,p)=p f'(p/q)=p-\sqrt{pq}$.

\begin{corollary}[Squared Hellinger gradient]
\label{cor:hellinger_gradient}
The gradient of the squared Hellinger divergence regularized score is:
\begin{equation}
\begin{aligned}
\nabla_x \mathcal{S}_{H^2}(x, y) &= \tau_1 \nabla_x f_y(x) - \tau_2 \sum_{i=1}^N p_{\tau_2}(i|x) \nabla_x f_i(x) \\
&\quad + \alpha \sum_{i=1}^N \sqrt{q_y(i)p(i|x)} \left[\nabla_x f_i(x) - \sum_{j=1}^N p(j|x) \nabla_x f_j(x)\right].
\end{aligned}
\label{eq:hellinger_gradient}
\end{equation}
\end{corollary}

Unlike KL-type divergences, the squared Hellinger distance is a true metric satisfying the triangle inequality. Its square root weighting $\sqrt{q_y(i)p(i|x)}$ provides more moderate gradient corrections compared to the logarithmic terms in KL divergences or the linear terms in JS divergence.

Empirical evaluation on ImageNet 128$\times$128 with ResNet-101 yielded FID = 2.15, (Table~\ref{tab:hellinger_results}). Interestingly, Hellinger achieves the \emph{highest precision} among all divergences, tied with FKL at 0.80, but exhibits the \emph{lowest recall} at 0.58. This precision-recall trade-off reveals Hellinger's implicit mode-seeking behavior despite being a symmetric metric.

The square root weighting creates a natural damping effect: when $p(i|x) \gg q_y(i)$, the gradient $\sqrt{q_y(i)p(i|x)}$ grows sublinearly, strongly discouraging the model from placing mass in low-target regions. Conversely, when $p(i|x) \ll q_y(i)$, the gradient also grows slowly, providing weak encouragement to explore under-represented modes. This \emph{bidirectional damping} results in conservative, high-fidelity generation that prioritizes precision over diversity.

While Hellinger outperforms the baseline (FID 2.19 $\to$ 2.15), it underperforms relative to RKL (2.14) and JS (2.13), and achieves the same recall as the baseline (0.58). This suggests that the bounded nature of Hellinger distance [always in $[0, 1]$ for probability distributions] limits its effectiveness for diffusion guidance, where stronger gradient signals are beneficial in early denoising steps. The metric's symmetric penalization of both under-coverage and over-extension may be suboptimal for conditional generation, where maintaining mode coverage is often more critical than preventing overshoot.

\begin{table}[h]
	\centering
    \vspace{-1em}
	\caption{Comparison of Squared Hellinger divergence result on ImageNet 128$\times$128 with 50k samples.}
	\vspace{5pt}
	\label{tab:hellinger_results}
	\begin{tabular}{lcccc}
		\toprule
		\textbf{Method} & \textbf{Classifier} &\textbf{FID} $\downarrow$ &  \textbf{Precision} $\uparrow$ & \textbf{Recall} $\uparrow$ \\
		\midrule
		\cite{ma2023elucidating} & ResNet-101 &  $\,2.19 $ &  $\,0.79 $&  $\,0.58 $\\
		Hellinger guided & ResNet-101  &  $\,2.15$ &  $\,0.80 $&  $\,0.58 $\\
		\bottomrule
	\end{tabular}
\end{table}

\section{Mathematical proofs}

\subsection{Proof of Proposition~\ref{pr1}}
\begin{proof}
The original guidance gradient is
\begin{align}
\nabla_x \log p_{\tau_1,\tau_2}(y|x) &= \tau_1 \nabla_x f_y(x) - \frac{\sum_{i=1}^N \exp(\tau_2 f_i(x)) \tau_2 \nabla_x f_i(x)}{\sum_{i=1}^N \exp(\tau_2 f_i(x))}\\
&= \tau_1 \nabla_x f_y(x) - \tau_2 \sum_{i=1}^N p_{\tau_2}(i|x) \nabla_x f_i(x).
\end{align}
For the entropy gradient 
\begin{align}
\nabla_x H(p(\cdot|x)) &= -\nabla_x \sum_{i=1}^N p(i|x) \log p(i|x)\\
&= -\sum_{i=1}^N \left[\nabla_x p(i|x) \log p(i|x) + p(i|x) \nabla_x \log p(i|x)\right].
\end{align}

Using the fact that $\nabla_x \log p(i|x) = (\nabla_x f_i(x) - \sum_{j} p(j|x) \nabla_x f_j(x))$, and $\nabla_x p(i|x) = p(i|x) \nabla_x \log p(i|x)$ we write
\begin{align}
\nabla_x H(p(\cdot|x)) = -\sum_{i=1}^N p(i|x)[\log p(i|x) + 1] \nabla_x \log p(i|x)
\end{align}
\begin{align}
\nabla_x H(p(\cdot|x)) = -\sum_{i=1}^N p(i|x)[\log p(i|x) + 1] \left[\nabla_x f_i(x) - \sum_{j=1}^N p(j|x) \nabla_x f_j(x)\right]
\end{align}
Combining them results to 
\begin{align}
\nabla_x \mathcal{S}_{\text{entropy}}(x, y) = -\nabla_x E_{\tau_1}(x, y) + \nabla_x E_{\tau_2}(x) + \lambda(t) \nabla_x H\!\left(p(\cdot|x)\right)
\end{align}
{\fontsize{8.5}{10}\selectfont
\begin{align}
= \tau_1 \nabla_x f_y(x) - \tau_2 \sum_{i=1}^N p_{\tau_2}(i|x) \nabla_x f_i(x) - \lambda(t) \sum_{i=1}^N p(i|x)[\log p(i|x) + 1] \left[\nabla_x f_i(x) - \sum_{j=1}^N p(j|x) \nabla_x f_j(x)\right].
\end{align}
}
\end{proof}
The additional $+1$ term in $[\log p(i|x) + 1]$ ensures proper normalization of the gradient, while the negative sign indicates that the gradient opposes concentration toward any single mode.
\subsection{Proof of Proposition~\ref{prop:general_f_div}}
\label{app:proof_general}

\begin{proof}
Let $D_f(q_y\|p)=\sum_{i} q_i\, f\!\left(\frac{p_i}{q_i}\right),
$ where $ p_i := p(i|x),\; q_i := q_y(i).$
Taking the gradient with respect to $x$, and noting that $q_i$ is constant:
\begin{align}
\nabla_x D_f
&= \sum_i q_i\, f'\!\left(\frac{p_i}{q_i}\right) \nabla_x\!\left(\frac{p_i}{q_i}\right) \nonumber\\
&= \sum_i f'\!\left(\frac{p_i}{q_i}\right) \nabla_x p_i.
\label{eq:pf1}
\end{align}
Using $\nabla_x p_i = p_i\,\nabla_x\log p_i$ and $\nabla_x \log p_i
= \nabla_x f_i(x) - \sum_j p_j\, \nabla_x f_j(x) =: g_i(x),$ we obtain
\begin{equation}
\nabla_x D_f
= \sum_i p_i\, f'\!\left(\frac{p_i}{q_i}\right) g_i(x)
= \sum_i w_f(q_i,p_i)\, g_i(x).
\label{eq:pf2}
\end{equation}
Combining with $\nabla_x \log p_{\tau_1,\tau_2}(y|x) = \tau_1 \nabla_x f_y(x) - \tau_2 \sum_i p_{\tau_2}(i|x)\,\nabla_x f_i(x)$
and the negative sign in $\mathcal{S}_{\mathcal{D}} = \log p_{\tau_1,\tau_2}(y|x) - \alpha D_f(q_y\|p)$
gives Proposition~\ref{prop:general_f_div}.
\end{proof}
\subsection{Proof of Corollary~\ref{cor:rkl_gradient}}
\label{app:proof_rkl}
\begin{proof}
Recall Proposition~2 gives
\begin{equation}
\nabla_x \mathcal{S}(x,y)
=
\tau_1 \nabla_x f_y(x)
-\tau_2 \sum_{i=1}^N p_{\tau_2}(i|x)\,\nabla_x f_i(x)
-\alpha \nabla_x D_f(q_y\|p(\cdot|x)),
\end{equation}
and for an $f$-divergence of the form $D_f(q\|p)=\sum_i q(i) f\!\left(\frac{p(i)}{q(i)}\right)$,
\begin{equation}
\nabla_x D_f(q_y\|p(\cdot|x))=\sum_{i=1}^N w_f\big(q_y(i),p(i|x)\big)\, g_i(x),
\qquad
w_f(q,p)=p\, f'\!\left(\frac{p}{q}\right),
\end{equation}
where $g_i(x)=\nabla_x \log p(i|x)$.

For reverse KL, $D_{\mathrm{KL}}(q_y\|p)=\sum_i q_y(i)\log\frac{q_y(i)}{p(i|x)}$
corresponds to $f(t)=-\log t$ with $t=\frac{p}{q}$. Hence $f'(t)=-1/t$ and
\begin{equation}
w_f\big(q_y(i),p(i|x)\big)
=
p(i|x)\, f'\!\left(\frac{p(i|x)}{q_y(i)}\right)
=
p(i|x)\left(-\frac{q_y(i)}{p(i|x)}\right)
=
-\,q_y(i).
\end{equation}
Therefore,
\begin{equation}
\nabla_x D_{\mathrm{KL}}(q_y\|p(\cdot|x))
=
\sum_{i=1}^N \big(-q_y(i)\big)\, g_i(x)
=
-\sum_{i=1}^N q_y(i)\, g_i(x),
\end{equation}
and the regularizer contribution in the score is
\(
-\alpha \nabla_x D_{\mathrm{KL}}(q_y\|p)=+\alpha \sum_i q_y(i) g_i(x).
\)
Substituting this into Proposition~2 yields Corollary~\ref{cor:rkl_gradient}.
\end{proof}

\subsection{Proof of Corollary~\ref{cor:fkl_gradient}}
\label{app:proof_fkl}
\begin{proof}
Forward KL is $D_{\mathrm{KL}}(p\|q_y)=\sum_i p(i|x)\log\frac{p(i|x)}{q_y(i)}$.
With our $f$-divergence convention $D_f(q\|p)=\sum_i q(i) f\!\left(\frac{p(i)}{q(i)}\right)$,
this corresponds to $f(t)=t\log t$ (again with $t=\frac{p}{q}$). Indeed,
\(
\sum_i q(i)\, t\log t = \sum_i p(i)\log\frac{p(i)}{q(i)}.
\)
Then $f'(t)=\log t + 1$, and the associated weight is
\begin{equation}
w_f\big(q_y(i),p(i|x)\big)
=
p(i|x)\Big(\log\frac{p(i|x)}{q_y(i)}+1\Big).
\end{equation}
Hence,
\begin{equation}
\nabla_x D_{\mathrm{KL}}(p\|q_y)
=
\sum_{i=1}^N p(i|x)\Big(\log\frac{p(i|x)}{q_y(i)}+1\Big)\, g_i(x),
\qquad g_i(x)=\nabla_x \log p(i|x),
\end{equation}
and plugging into Proposition~2 gives the regularizer term
\(
-\alpha \nabla_x D_{\mathrm{KL}}(p\|q_y)
=
-\alpha \sum_i p(i|x)\big(\log\frac{p(i|x)}{q_y(i)}+1\big) g_i(x),
\)
which is exactly Corollary~\ref{cor:fkl_gradient}.
\end{proof}

\subsection{Proof of Corollary~\ref{cor:js_gradient}}
\label{app:proof_js}
\begin{proof}
The Jensen--Shannon divergence is
\begin{equation}
D_{\mathrm{JS}}(q_y\|p)
=
\frac{1}{2}D_{\mathrm{KL}}(q_y\|m)+\frac{1}{2}D_{\mathrm{KL}}(p\|m),
\qquad
m(i)=\frac{q_y(i)+p(i|x)}{2}.
\end{equation}
Since $q_y$ is fixed, only $p(\cdot|x)$ (and thus $m$) depends on $x$.
Differentiate $D_{\mathrm{JS}}$ w.r.t.\ $p(i|x)$ (treating $p$ as free variables first).
A direct computation gives the partial derivative
\begin{equation}
\frac{\partial}{\partial p(i)} D_{\mathrm{JS}}(q_y\|p)
=
\frac{1}{2}\log\frac{p(i)}{m(i)}.
\end{equation}
(Indeed, the $q_y\|m$ term contributes $-\frac{q_y(i)}{4m(i)}$,
the $p\|m$ term contributes $\frac{1}{2}\big(\log\frac{p(i)}{m(i)}+1\big)-\frac{p(i)}{4m(i)}$,
and the rational terms cancel because $q_y(i)+p(i)=2m(i)$.)

Therefore, by the chain rule,
\begin{equation}
\nabla_x D_{\mathrm{JS}}(q_y\|p(\cdot|x))
=
\sum_{i=1}^N \frac{1}{2}\log\frac{p(i|x)}{m(i)}\, \nabla_x p(i|x)
=
\sum_{i=1}^N \frac{p(i|x)}{2}\log\frac{p(i|x)}{m(i)}\, g_i(x),
\end{equation}
where $\nabla_x p(i|x)=p(i|x)g_i(x)$ and $g_i(x)=\nabla_x\log p(i|x)$.
Since $\frac{p(i|x)}{m(i)}=\frac{2p(i|x)}{p(i|x)+q_y(i)}$, this can be written as
\begin{equation}
\nabla_x D_{\mathrm{JS}}(q_y\|p(\cdot|x))
=
\sum_{i=1}^N \frac{p(i|x)}{2}\log\frac{2p(i|x)}{p(i|x)+q_y(i)}\, g_i(x).
\end{equation}
Substituting into Proposition~2 and applying the $-\alpha$ factor in the score yields Corollary~\ref{cor:js_gradient}.
\end{proof}

\subsection{Proof of Lemma~\ref{lem:rkl_decomposition}}
\label{app:proof_lemma}
\begin{proof}
By definition,
\begin{equation}
D_{\mathrm{KL}}(q_y\|p(\cdot|x))=\sum_{i=1}^N q_y(i)\log\frac{q_y(i)}{p(i|x)}
=
\text{const} - \sum_{i=1}^N q_y(i)\log p(i|x),
\end{equation}
so
\begin{equation}
-\nabla_x D_{\mathrm{KL}}(q_y\|p(\cdot|x))=\sum_{i=1}^N q_y(i)\,\nabla_x \log p(i|x).
\end{equation}
For softmax probabilities $p(i|x)=\frac{e^{f_i(x)}}{\sum_k e^{f_k(x)}}$,
\begin{equation}
\nabla_x \log p(i|x)=\nabla_x f_i(x)-\sum_{j=1}^N p(j|x)\nabla_x f_j(x).
\end{equation}
Multiplying by $q_y(i)$ and summing over $i$ gives
\begin{align}
-\nabla_x D_{\mathrm{KL}}(q_y\|p(\cdot|x))
&=
\sum_{i=1}^N q_y(i)\nabla_x f_i(x)
-
\Big(\sum_{i=1}^N q_y(i)\Big)\sum_{j=1}^N p(j|x)\nabla_x f_j(x) \nonumber\\
&=
\sum_{i=1}^N q_y(i)\nabla_x f_i(x)
-
\sum_{j=1}^N p(j|x)\nabla_x f_j(x),
\end{align}
since $\sum_i q_y(i)=1$. This is the claimed decomposition.
\end{proof}

\subsection{Proof of Proposition~\ref{prop:gaussian_rkl}}
\label{app:proof_gaussian}
\begin{proof}
Let $\ell_k(x)=\log b_k+\log f_k(x)$. For Gaussian components, $
\nabla_x \ell_k(x)=\nabla_x \log f_k(x)=\Sigma_k^{-1}(\mu_k-x)$. Additionally, since $\nabla_x \log f_k(x) = \frac{1}{f_k(x)} \nabla_x f_k(x) \rightarrow \nabla_x f_k(x) = f_k(x) \nabla_x \log f_k(x) $. Substituting these into Corollary~\ref{cor:rkl_gradient} gives
{\fontsize{8.5}{10}\selectfont
\begin{equation}
\begin{aligned}
\nabla_x \mathcal{S}_{\text{RKL}}(x, y)
&= \tau_1 f_y(x) \nabla_x \ell_y(x) - \tau_2 \sum_{k=1}^K p_{\tau_2}(k|x) f_k(x) \nabla_x \ell_k(x) \\
&\quad + \alpha \sum_{k=1}^K q_y(k) \Big(  f_k(x) \nabla_x \ell_k(x) - \sum_{j=1}^K  f_j(x)  w_j(x)\nabla_x \ell_j(x)\Big) \\
&= \sum_{k=1}^K  f_k(x) \Sigma_k^{-1}(\mu_k-x)\\ &\times\Big[\tau_1\mathbb{I}_{k=y}-\tau_2 p_{\tau_2}(k|x)+\alpha q_y(k)\Big] \\ &
     \;-\;\alpha\Big(\sum_{k=1}^K q_y(k)\Big)\sum_{j=1}^K f_j(x) w_j(x)\Sigma_j^{-1}(\mu_j-x).
\end{aligned}
\end{equation}}
Since $\sum_k q_y(k)=1$, collecting the coefficient of each $ f_k(x) \Sigma_k^{-1}(\mu_k-x)$ yields
$\Gamma_k(x,y,\alpha,\epsilon)=\tau_1\mathbb{I}_{k=y}-\tau_2 p_{\tau_2}(k|x)+\alpha(q_y(k)-w_k(x))$ as stated.
\end{proof}

\subsection{Proof of Corollary~\ref{cor:hellinger_gradient}}
\label{app:proof_hellinger}
\begin{proof}
The squared Hellinger distance can be written as
\begin{equation}
D_{H^2}(q_y\|p(\cdot|x))
=
\sum_{i=1}^N \big(\sqrt{q_y(i)}-\sqrt{p(i|x)}\big)^2
=
\sum_i q_y(i) + \sum_i p(i|x) - 2\sum_i \sqrt{q_y(i)p(i|x)}.
\end{equation}
Since $\sum_i q_y(i)=1$ and $\sum_i p(i|x)=1$, this reduces to
\(
D_{H^2}(q_y\|p)=2-2\sum_i \sqrt{q_y(i)p(i|x)}.
\)
Thus,
\begin{equation}
\nabla_x D_{H^2}(q_y\|p(\cdot|x))
=
-2\sum_{i=1}^N \nabla_x \sqrt{q_y(i)p(i|x)}
=
-2\sum_{i=1}^N \frac{\sqrt{q_y(i)}}{2\sqrt{p(i|x)}}\,\nabla_x p(i|x).
\end{equation}
Using $\nabla_x p(i|x)=p(i|x)\,g_i(x)$ with $g_i(x)=\nabla_x\log p(i|x)$,
\begin{equation}
\nabla_x D_{H^2}(q_y\|p(\cdot|x))
=
-\sum_{i=1}^N \sqrt{q_y(i)p(i|x)}\, g_i(x).
\end{equation}
Finally, for the score
\(
\mathcal{S}_{H^2}(x,y)=\log p_{\tau_1,\tau_2}(y|x)-\alpha D_{H^2}(q_y\|p(\cdot|x)),
\)
we obtain
\begin{equation}
\nabla_x \mathcal{S}_{H^2}(x,y)
=
\tau_1 \nabla_x f_y(x)
-\tau_2 \sum_{i=1}^N p_{\tau_2}(i|x)\nabla_x f_i(x)
+\alpha \sum_{i=1}^N \sqrt{q_y(i)p(i|x)}\, g_i(x),
\end{equation}
which is the stated corollary (after rewriting $g_i(x)$ via the softmax identity if desired).
\end{proof}

\section{Implementation details} \label{imp}
\subsection{Fine-Tuning with ECE Smooth Regularization} \label{eceimp}

To improve the calibration of the classifier without compromising its predictive performance, we incorporate the ECE Smooth loss as a regularization term alongside the standard cross-entropy objective. Specifically, during fine-tuning, the total loss combines the classification loss with the ECE Smooth term. The following algorithm outlines the training procedure used in our implementation

\begin{algorithm}
\caption{ECE Smooth Fine-Tuning}
\label{alg:ece-smooth-finetune}
\begin{algorithmic}[1]
\Require Number of epochs $E$, batch size $B$, regularization weight $\lambda$, classifier $\mathcal{C}_\phi$, smoothing constant $\beta$
\For{$e = 1, \ldots, E$}
    \For{each batch $\{(x^{(i)}, y^{(i)})\}_{i=1}^B$}
        \State $\hat{p}^{(i)} \leftarrow \mathcal{C}_\phi(x^{(i)})$ \Comment{Predicted probability vector}
        \State $a^{(i)} \leftarrow \mathbb{1}\left( \arg\max_j \hat{p}^{(i)}_j = y^{(i)} \right)$ \Comment{Correctness indicator}
        \State Compute classification loss: 
        \State \hspace{\algorithmicindent} $\mathcal{L}_{\text{CE}} = \frac{1}{B} \sum_{i=1}^B -\log \hat{p}^{(i)}_{y^{(i)}}$
        \State Compute ECE smooth regularizer:
        \State \hspace{\algorithmicindent} $\mathcal{L}_{\mathrm{ECE}} = \frac{1}{B} \sum_{i=1}^{B} \sqrt{ \left( \max_j \hat{p}^{(i)}_j - a^{(i)} \right)^2 + \beta }$
        \State Combine losses: $\mathcal{L} = \mathcal{L}_{\text{CE}} + \lambda \cdot \mathcal{L}_{\mathrm{ECE}}$
        \State Update $\phi$ using gradient of $\mathcal{L}$
    \EndFor
\EndFor
\State \Return fine-tuned classifier $\mathcal{C}_\phi$
\end{algorithmic}
\end{algorithm}
We fine-tuned the official \cite{dhariwal2021diffusion} classifier from \href{https://openaipublic.blob.core.windows.net/diffusion/jul-2021/128x128_classifier.pt}{here} using ECE Smooth regularizer with a regularization weight of $\lambda = 1$ and a smoothing constant of $\beta = 0.0001$. The training was performed for a total of $12000$ iterations with $B=256$ in our setup. We used a single NVIDIA H100 80 GB and the official ILSVRC2012 dataset for fine-tuning.

\subsection{Sampling}
The off-the-shelf classifiers are the official Pytorch checkpoints at \href{https://download.pytorch.org/models/resnet50-11ad3fa6.pth}{here}. We used the $(\gamma_t)$ schedule and other default settings from \cite{ma2023elucidating} to ensure a fair comparison, and incorporated our proposed methods on top of this baseline to assess their improvements. We used the diffusion model from \cite{dhariwal2021diffusion} without retraining. Through all the experiments, we evaluated the FID score using a batch size of $B=256$, and the reference dataset batches contain pre-computed statistics over the whole dataset provided \href{https://github.com/openai/guided-diffusion/tree/main/evaluations}{here}. We also used a single NVIDIA A100 SXM4 40 GB for this stage. 

The divergence regularization adds negligible computational overhead. All methods share $\mathcal{O}(B \cdot C)$ complexity with the baseline, where divergence computation (one softmax + one divergence term) adds 50-80\% cost to gradient computation. Since classifier gradients represent a small fraction of total sampling time (dominated by diffusion model forward passes), end-to-end overhead is less than 3\% on ImageNet 128$\times$128 with 250 DDPM steps. Memory overhead is also minimal ($<$12 MB for batch size 256).

\section{Additional experiments} \label{addexp}

\subsection{Tilted sampling}
Table~\ref{tab:fid_t} presents an ablation study of the TERM tilt parameter $t$ for classifier-guided diffusion sampling. The results demonstrate a non-monotonic relationship between $t$ and generation quality measured by FID. Optimal performance is achieved at $t = -0.2$ (FID = 5.28), improving upon the standard averaging baseline at $t = 0$ (FID = 5.34). Performance degrades at both extremes: positive tilting ($t = 0.1$) yields FID = 5.45, while increasingly negative values deteriorate from FID = 5.30 at $t = -0.1$ to FID = 5.63 at $t = -0.5$.

The optimal performance at mild negative tilting ($t = -0.2$) reveals that classifier guidance benefits from down-weighting high-confidence predictions. When $t < 0$, TERM effectively implements confidence tempering by emphasizing lower-probability samples, preventing the diffusion process from over-optimizing toward potentially spurious classifier modes. This acts as an implicit regularizer that reduces sensitivity to classifier overconfidence and biases. In contrast, positive tilting amplifies already-dominant high-confidence predictions, leading to less diverse generations and potential adversarial artifacts. The degradation at strongly negative values ($t = -0.5$) suggests that excessive emphasis on low-confidence predictions provides overly weak guidance signals. Thus, the sweet spot at $t = -0.2$ balances leveraging classifier knowledge while avoiding over-reliance on its most confident predictions, resulting in more stable and higher-quality image generation.

\begin{table}[h]
  \centering
  \caption{ Ablation study of $t$ in tilted guidance with respect to FID. The classifier is the
ResNet-50. The diffusion model is from \cite{dhariwal2021diffusion}. We generate 10k
ImageNet $128\times128$ samples with 250 DDPM steps for evaluation.
}
\vspace{5pt}
  \label{tab:fid_t}
  \begin{tabular}{lcccccc}
    \toprule
    $t$ & 0.1 & 0 & $-0.1$ & $-0.2$ & $-0.3$ & $-0.5$\\
    \midrule
    FID      & 5.45 & 5.34 & 5.30 & 5.28 & 5.40 & 5.63\\
    \bottomrule
  \end{tabular}
\end{table}
\subsection{Entropy guidance sampling }
The hyperparameter here is the adaptive entropy regularization weight $\lambda_t$. We first tested a constant value $\lambda_t = 0.1$, which resulted in almost the same FID as the baseline. Therefore, we selected an adaptive approach for the weight as $\lambda_t \in \left[ 0.05, 0.2\right]$. This will start from the greatest value at the start during early denoising steps when the image is heavily corrupted and classifier predictions are unreliable, then gradually decreases to the minimum value. This time-dependent approach aligns with the inherent dynamics of the diffusion process; early steps benefit from exploration (high entropy) while later steps require precision (low entropy) to preserve fine-grained class characteristics. This resulted in the best FID, which was reported in the experiments. 
\subsection{Divergence guidance sampling } \label{addexdiv}
The selection of the target distribution $q_y(\cdot)$ in Equation~\eqref{eq:f_divergence_general} is critical for balancing class fidelity with mode coverage. A complete uniform distribution $q_y(i) = 1/N$ would be class-agnostic and detrimental to conditional generation, while a one-hot distribution $q_y(i) = \mathbb{I}_{i=y}$ would eliminate the regularization's diversity-preserving effect. We adopt a parameterized target distribution:
\begin{equation}
q_y(i) = (1-\epsilon)\frac{1}{N} + \epsilon \:\mathbb{I}_{i=y},
\label{eq:target_distribution}
\end{equation}
where $\epsilon \in [0, 1]$ controls the bias toward the target class $y$. This formulation:
\begin{itemize}
    \item At $\epsilon = 0$: reduces to uniform, maximizing diversity but losing class conditioning
    \item At $\epsilon = 1$: reduces to one-hot, maximizing class fidelity but losing regularization
    \item At $\epsilon \in (0, 1)$: balances target class emphasis with exploration of adjacent modes
\end{itemize}
Based on our experiments (see Table~\ref{tab:fid_alpha}), we selected $\epsilon = 0.1$. Table~\ref{tab:fid_alpha} demonstrates the FID analysis for the weight parameters.   
\begin{table}[h]
  \centering
  \caption{ Ablation study of $\alpha$ in divergence guidance with respect to FID. The classifier is the
ResNet-50. The diffusion model is from \cite{dhariwal2021diffusion}. We generate 10k
ImageNet $128\times128$ samples with 250 DDPM steps for evaluation.}
\vspace{5pt}
  \label{tab:fid_alpha}
  \begin{tabular}{lcccccccc}
    \toprule
    $\alpha$ & 0.0 & 0.05 & 0.1 & 0.15 & 0.1 & 0.17&0.1\\
    \midrule
        $\epsilon$ & 0.0 & 0.1 & 0.1 & 0.1 & 0.05 & 0.08&0.2\\
    \midrule
    FID      & 5.34 & 5.31 & 5.12 & 5.29 & 5.37 & 5.22&5.41\\
    \bottomrule
  \end{tabular}
\end{table}

Thus, we selected $\alpha = 0.1$ for our experiments on Resnet-50. We also evaluated a similar study for JS and forward KL divergences using Resnet-101. Table~\ref{tab:fid_alpha2} demonstrates the FID analysis for the weight parameter with $\epsilon = 0.1$.   

\begin{figure}
    \centering
    \begin{subfigure}{0.92\linewidth}
        \centering
        \includegraphics[width=\linewidth]{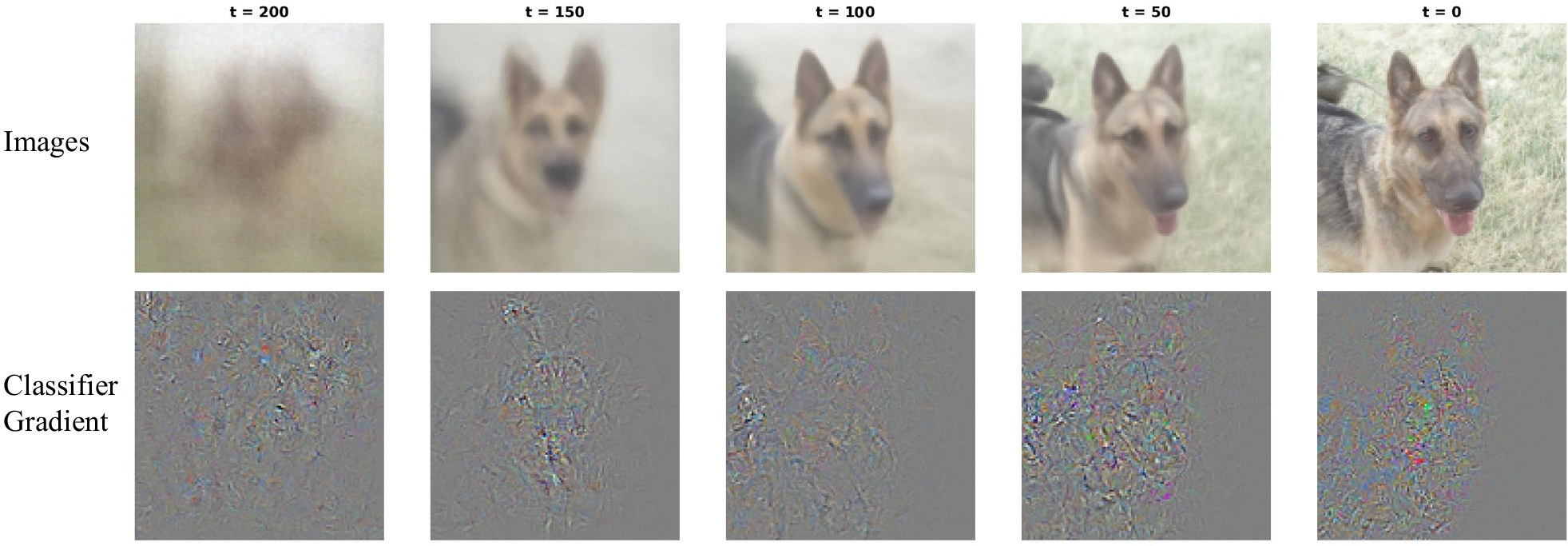}
        \caption{Baseline ($\tau_1 = \tau_2 = 1$)}
        \label{fig:cmp-baseline1}
    \end{subfigure}

    \vspace{0.5em} 

    \begin{subfigure}{0.92\linewidth}
        \centering
        \includegraphics[width=\linewidth]{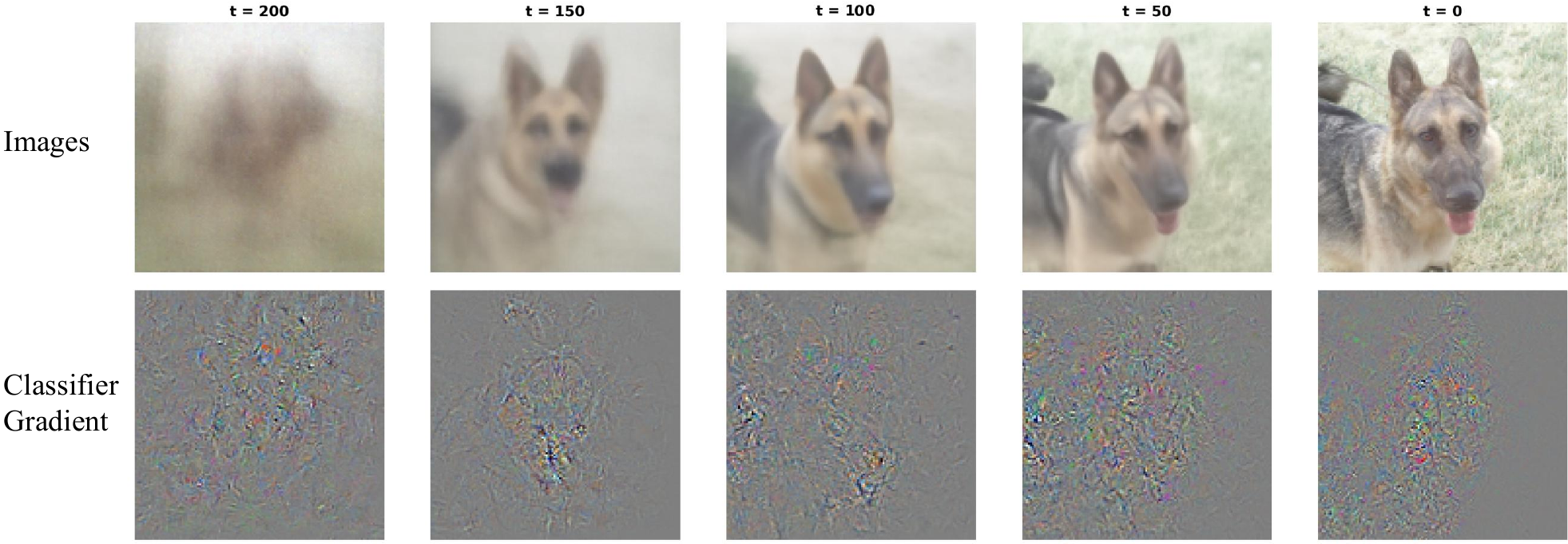}
        \caption{Baseline ($\tau_1 = 1$ and  $\tau_2 = 0.5$) \cite{ma2023elucidating}}
        \label{fig:cmp-baseline2}
    \end{subfigure}

    \vspace{0.5em}

    \begin{subfigure}{0.92\linewidth}
        \centering
        \includegraphics[width=\linewidth]{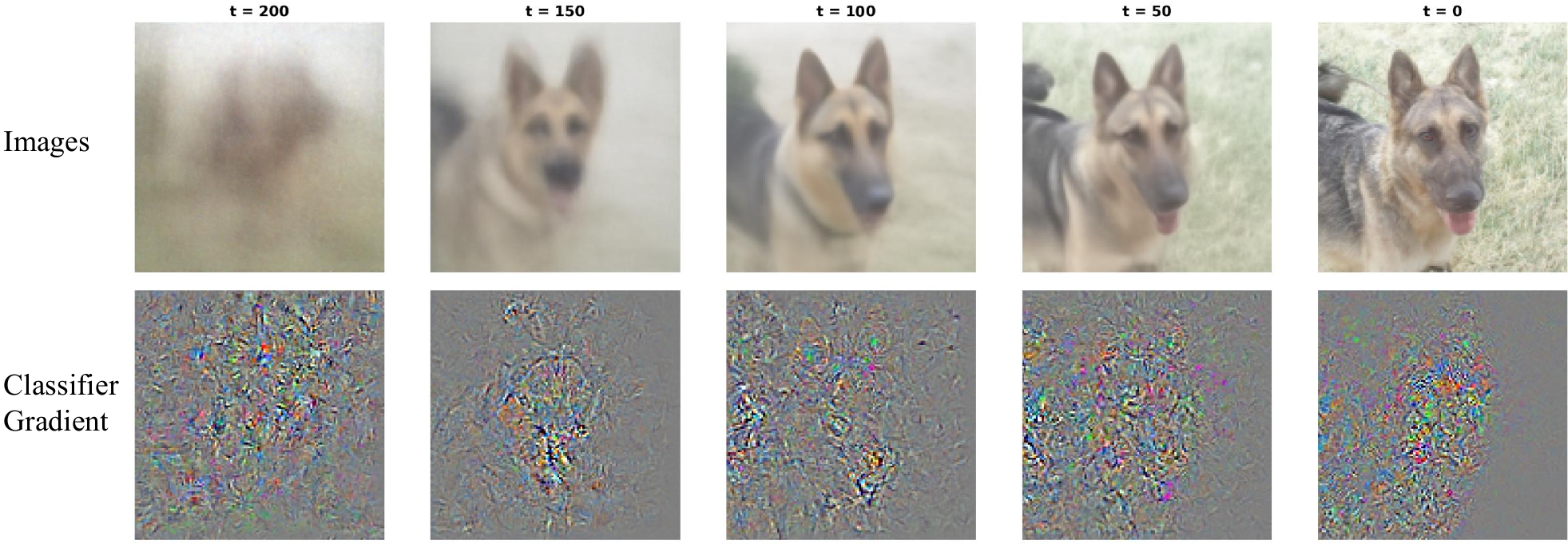}
        \caption{Reverse KL divergence regularization}
        \label{fig:cmp-fdiv}
    \end{subfigure}

    \caption{The visual comparison of intermediate sampling pictures and classifier gradient figures. The seed is fixed for
direct comparison.}
    \label{fig:cmp}
\end{figure}

\begin{table}[h]
  \centering
  \caption{ Ablation study of $\alpha$ in divergence guidance with respect to FID. The classifier is the
ResNet-101. The diffusion model is from \cite{dhariwal2021diffusion}. We generate 50k
ImageNet $128\times128$ samples with 250 DDPM steps for evaluation.}
\vspace{5pt}
  \label{tab:fid_alpha2}
  \begin{tabular}{lccccccc}
    \toprule
    $\alpha$ & 0.1 & 0.09 & 0.08 & 0.1 & 0.09 & 0.08\\
    \midrule
        Divergence & JS &JS &JS & FKL& FKL& FKL\\
    \midrule
    FID      & 2.16&2.13&2.17&2.20&2.16&2.18\\
    \bottomrule
  \end{tabular}
\end{table}
Therefore, $\alpha = 0.09$ was the optimal value for our experiments using Resnet-101. Figure~\ref{fig:cmp} provides a visual comparison of classifier gradient behaviors across three guidance strategies during the diffusion sampling process, with differences in both the gradient evolution and final image quality. The visualization tracks the evolution from heavily noised samples (t=200) to clean images (t=0), revealing distinct patterns in how each method maintains gradient activity and ultimately affects the generated output. In the baseline approach \ref{fig:cmp-baseline1} with equal temperatures, the classifier gradients exhibit rapid concentration. The improved baseline \ref{fig:cmp-baseline2} shows marginally better gradient preservation but still suffers from premature collapse in the middle stages (t=100 to t=50), producing a final image with comparable quality but subtle differences in texture rendering. In contrast, our reverse KL divergence regularization method \ref{fig:cmp-fdiv} demonstrates persistent gradient activity throughout the entire sampling trajectory, culminating in a final image with noticeably sharper details, better color saturation, and more natural texture—particularly visible in the dog's fur and facial features. The gradient maps maintain rich, distributed patterns even in later denoising steps, with visible activity across multiple spatial regions rather than collapsing to isolated points. This sustained gradient diversity validates our theoretical analysis.


\end{document}